\documentclass[10pt,journal,compsoc]{IEEEtran}
\ifCLASSOPTIONcompsoc
  \usepackage[nocompress]{cite}
\else
  \usepackage{cite}
\fi
\usepackage{graphicx}
\usepackage{amsmath}
\usepackage{amssymb}
\usepackage{booktabs}

\usepackage{algorithm}
\usepackage{alltt}
\usepackage{capt-of}
\usepackage{times}
\usepackage{epsfig}
\usepackage{graphicx}
\usepackage{amsmath}
\usepackage{amssymb}
\usepackage{blindtext}
\usepackage{overpic}
\usepackage{transparent}
\usepackage{booktabs}
\usepackage{multirow}
\usepackage[super]{nth}
\usepackage[format=plain,labelformat=simple,labelsep=period,font=small,skip=4pt,compatibility=false]{caption}
\usepackage[font=footnotesize,skip=2pt]{subcaption}
\usepackage{makecell}
\usepackage[colorlinks,linkcolor=blue]{hyperref}
\usepackage[dvipsnames,table,xcdraw]{xcolor}

\makeatletter
\newcommand\notsotiny{\@setfontsize\notsotiny\@vipt\@viipt}

\makeatother

\usepackage{etoolbox}
\usepackage[binary-units]{siunitx}
\robustify\bfseries
\DeclareSIUnit{\inch}{inch}

\newcolumntype{C}[1]{>{\centering\arraybackslash}p{#1}}

\usepackage{xspace}
\makeatletter
\DeclareRobustCommand\onedot{\futurelet\@let@token\@onedot}
\def\@onedot{\ifx\@let@token.\else.\null\fi\xspace}
\def\eg{\emph{e.g}\onedot} 
\def\ie{\emph{i.e}\onedot} 
 
\def\etc{\emph{etc}\onedot} 
 
\def\etal{\emph{et al}\onedot}
\makeatother

\usepackage{pifont}
\definecolor{lightgray}{rgb}{0.9, 0.9, 0.9}
\definecolor{lgray}{rgb}{0.66, 0.66, 0.66}

\newcommand{\cb}{\texttt{Combo}}
\ifCLASSINFOpdf
\else
\fi
\begin{document}
% \title{\includegraphics[width=0.07\linewidth]{fig/pie.pdf}MotionPIE: Ra\textcolor{red}{P}idly Crafting the Holist\textcolor{red}{I}c Speech-Driven Human Motion as You Desir\textcolor{red}{E}}
% \title{\includegraphics[width=0.05\linewidth]{fig/combo.png} \texttt{Combo}: \textbf{\texttt{Co}}-speech holistic 3D human \textbf{\texttt{m}}otion \\ generation and efficient customiza\textbf{\texttt{b}}le adaption in harm\textbf{\texttt{o}}ny}
\title{\textbf{\texttt{Combo}}: \textbf{\texttt{Co}}-speech holistic 3D human \textbf{\texttt{m}}otion \\ generation and efficient customiza\textbf{\texttt{b}}le adaptation in harm\textbf{\texttt{o}}ny}
% \title{MotionPIE: Rapidly Crafting the Holistic Speech-Driven Human Motion as You Desire}

\author{Chao~Xu,
        Mingze~Sun,
        Zhi-Qi~Cheng,
        Fei~Wang,
        Yang~Liu,
        Baigui~Sun,
        Ruqi~Huang,
        Alexander~Hauptmann% <-this % stops a space
\IEEEcompsocitemizethanks{
\IEEEcompsocthanksitem C. Xu, Y. Liu and B. Sun are with Wolf 1069B Lab, Sany Group (chaoxuxc@gmail.com, sunbaigui85@gmail.com).
\IEEEcompsocthanksitem M. Sun, and R. Huang are with Tsinghua Shenzhen International Graduate School, Tsinghua University (e-mail: smz22@mails.tsinghua.edu.cn, ruqihuang@sz.tsinghua.edu.cn).
\IEEEcompsocthanksitem Z.-Q. Cheng is with the School of Engineering and Technology, University of Washington, Tacoma, WA, USA (e-mail: zhiqics@uw.edu).
\IEEEcompsocthanksitem F. Wang is with Alibaba Group (stevenwang19882018@gmail.com).
\IEEEcompsocthanksitem Y. Liu is with the Department of Engineering, King’s College
London (e-mail: yang.15.liu@kcl.ac.uk)
\IEEEcompsocthanksitem A. Hauptmann is with the Language Technologies Institute, Carnegie Mellon University, Pittsburgh, PA, USA (e-mail: alex@cs.cmu.edu).
\IEEEcompsocthanksitem C. Xu, M. Sun, and Z.-Q. Cheng contributed equally; author order is random. B. Sun and R. Huang are the corresponding authors.
}
\thanks{Manuscript received April 19, 2005; revised August 26, 2015.}}


\markboth{Journal of \LaTeX\ Class Files,~Vol.~14, No.~8, August~2015}%
{Shell \MakeLowercase{\textit{et al.}}: Bare Demo of IEEEtran.cls for Computer Society Journals}



\IEEEtitleabstractindextext{%
\begin{abstract}

In this paper, we propose a novel framework, \cb, for harmonious co-speech holistic 3D human motion generation and efficient customizable adaption. 
In particular, we identify that one fundamental challenge as the multiple-input-multiple-output (MIMO) nature of the generative model of interest. 
More concretely, on the input end, the model typically consumes both speech signals and character guidance (\emph{e.g., }identity and emotion), which hinders further adaptation to varying guidance; on the output end, holistic human motions mainly consist of facial expressions and body movements, which are inherently correlated but non-trivial to coordinate in current data-driven generation process. 
In response to the above challenge, we propose tailored designs to both ends.  
For the former, we propose to pre-train on data regarding a fixed identity with neutral emotion, and defer the incorporation of customizable conditions (identity and emotion) to fine-tuning stage, which is boosted by our novel \texttt{X}-Adapter for parameter-efficient fine-tuning. 
For the latter, we propose a simple yet effective transformer design, DU-Trans, which first divides into two branches to learn individual features of face expression and body movements, and then unites those to learn a joint bi-directional distribution and directly predicts combined coefficients. 
Evaluated on BEAT2 and SHOW datasets, \texttt{Combo} is highly effective in generating high-quality motions but also efficient in transferring identity and emotion.
Project website: \href{https://xc-csc101.github.io/combo/}{Combo}.

\end{abstract}

}

\maketitle

\IEEEdisplaynontitleabstractindextext

% \fi
\IEEEpeerreviewmaketitle

\section{Introduction}
\begin{figure*}[t!]
    \centering
	\includegraphics[width=0.85\textwidth]{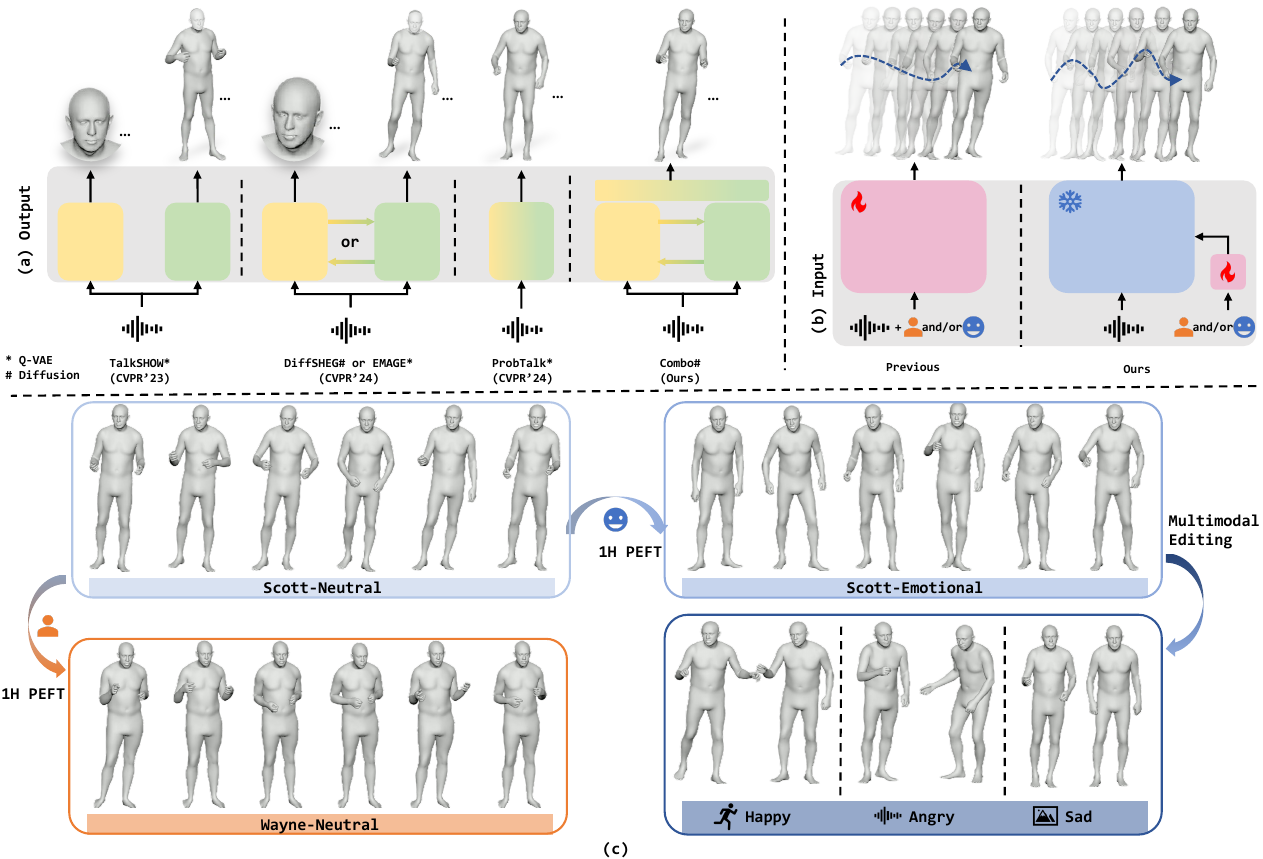}
	\caption{\textbf{An illustrative comparison of SOTA methods and our approach.} (a) \textbf{On output end}, TalkSHOW~\cite{yi2023generating} adopts two separate networks to predict audio-synchronized expressions and gestures, thus leading to a disjointed issue. Subsequent work~\cite{chen2024diffsheg, liu2024emage} only account for bidirectional interaction and use two heads predict independently, still leading to a risk of disharmony. ProbTalk~\cite{liu2024towards} directly learns holistic motion for harmony but the specificity of each component is lost. Conversely, our method not only achieves \emph{holistic harmony} but also maintains \emph{individual uniqueness}. (b) \textbf{On input end}, previous works typically learn a comprehensive model to fit the multiple inputs, including audio signals and other conditions. Thus they require training or finetuning the entire network for different conditions. In contrast, we pre-train a model in a relatively clear setting, and adapt various conditions in the fine-tuning stage. Our design allows flexible and efficient adaptation (\emph{1 Hour PEFT}) for identity customization and emotion transfer, as depicted in part (c). Notably, we only utilize the embeddings from emotional text prompts encoded by CLIP as guidance during training. At inference, other modalities such as motion clips, audios, and images can all serve as emotional conditions, thereby supporting flexible multi-modal editing, as depicted in part (c).}
	\label{fig:teaser}
\end{figure*}

In this paper, we study the problem of co-speech holistic 3D human motions generation~\cite{yi2023generating}, namely, given speech signal and character conditions (\emph{e.g., }identity, emotion), generating facial expressions and body movements including hand gestures and body motions. 
This generation task is crucial in crafting digital avatars as it significantly enhances the interaction informativeness and vividness of the latter~\cite{chen2024diffsheg, liu2024emage}, therefore attracting increasing interest from generative AI research. 

One fundamental challenge of this task stems from its multi-input-multi-output (MIMO) nature: 
1) the input end gathers not only various factors including speech contents, rhythms, and semantics, but also character conditions like identities and emotions; 
2) the output end delivers both facial and body motions, which are inherently related but non-trivial to align. 
For humans, such a perplexing system is well governed by a nervous system, yielding harmonious, adaptive (\emph{e.g., }regarding emotion changes) holistic motions in real life. 
On the other hand, generative models typically take a data-driven approach to learning from human videos. 
While being straightforward, it is highly non-trivial to learn the inherent relationships within the MIMO system purely from data.

In light of the above, 
we examine the holistic human motion generation task from a principled perspective and advocate two novel designs tailored for improving data-driven generative model. 
Our key insight is to alleviate modeling complexity in both input and output end, leading to a \emph{harmonious but also adaptive} generative network. 
For the input end, instead of training an encoder that consumes all at once, we pre-train a model in a relatively clear setting, and defer the injection of important conditions to the fine-tuning stage; 
For the output end, we propose a divide-and-unite strategy to address the trade-off between learning accurate facial and body movements and guaranteeing coherence.
In below, we motivate and explain the above designs in more detail. 

Regarding the \textbf{multiple input end}, our generative model is expected to take speeches as generation guidance, which contain factors such as contents, rhythms, semantics, \etc. 
It is worth noting that the same speech guidance can lead to different motion sequences with respect to varying conditions (\emph{e.g., }identity and emotion), complicating the learning process. 
The significant contributions to high-quality dataset construction~\cite{liu2022beat, liu2024emage}, containing rich annotations as well as great variability in identity and emotions, offer good opportunities for learning a comprehensive model. 
However, we argue that learning a comprehensive model lacks flexibility in further adaptation.
In fact, generative models have been widely desired to be controllable and customizable~\cite{chen2024anydoor, li2024photomaker, xu2024xagen, peng2024portraitbooth}. 
Regarding holistic human motion generation, one would naturally expect a flexible, adaptive digital avatar with respect to changes in both short-term emotional states (emotion) and long-term personality (identity). 
Unfortunately, recent efforts~\cite{yi2023generating, liu2024emage, chen2024diffsheg, yin2023emog, qi2024emotiongesture} require an extensive training or full-parameter fine-tuning process to accommodate to the emotional data or newly introduced identities, a limitation that is particularly exacerbated in VQ-VAE-based methods~\cite{yi2023generating, liu2024emage}, as shown in Fig.~\ref{fig:teaser}(b).

Motivated by the above observations, we take a pre-train-and-fine-tune approach. 
During pre-training, we train a model with data from a fixed identity in neutral emotion. 
The injection of specific conditions, such as alternative identities and/or emotions, is deferred to the fine-tuning stage. 
Consequently, this technique overcomes the complexity of training a universal model with various conditions by a two-stage scheme, \ie, first train on a simple condition to establish a basic model, and then perform fast fine-tuning to adapt this model to different scenarios.
Our key design in this part is a novel plug-and-play adapter for Parameter-Efficient Fine-Tuning (PEFT)~\cite{han2024parameter}, \texttt{X}-Adapter, which is tailored to our pipeline and allows for effective and efficient fine-tuning on emotion transfer and personalized generation. 
Specifically, our treatment on input end admits several advantages: 1) It eliminates complex conditions by training a simple basic model, which can be trained easier and more through for improved performance; 2) Leveraging the advanced pre-training, the fast fine-tuning phase requires only minimal parameter adjustments to guide the model in the desired conditional direction while maintaining excellent performance; 3) As a by-product, it further enables versatile editing during inference benefiting from flexible \texttt{X} cues, which is a capability that previous approaches~\cite{yi2023generating, chen2024diffsheg, liu2024emage} could not attain. For emotion, we can employ multi-modal conditions within the CLIP domain~\cite{radford2021learning} to indicate the short-term emotion, which supports flexible and generalizable zero-shot editing. 
For identity, we define identity codes as several interpretable statistics derived from the face and body parameters of SMPLX~\cite{pavlakos2019expressive} to depict long-term personality.

Regarding the \textbf{multiple output end}, holistic human motion involves both facial expressions and body movements with different patterns~\cite{chen2024diffsheg, liu2024emage, ng2024audio}, presents a dual challenge: direct holistic modeling~\cite{liu2024towards} is difficult to maintain uniqueness, while separate modeling struggles to generate harmonious full-body natural movements.  
For the former, ProbTalk~\cite{liu2024towards} directly models the holistic motions for coordination but fails to learn accurate individual distribution.
For the latter, TalkSHOW~\cite{yi2023generating} completely ignores the interconnection and results in disjointed coordination among different motion components. 
Subsequent works~\cite{liu2024emage, chen2024diffsheg} identify this discrepancy and explicitly utilize a unidirectional flow between face and body to enhance the correlation. 
However, the current arts are inadequate to achieve holistic harmony yet individual distinctive, leaving sufficient room for improvement, as shown in Fig.~\ref{fig:teaser}(a). 

To this end, we propose a simple yet effective transformer design, \textbf{DU}-Trans, which first \textbf{D}ivides into two branches to learn individual features of face expression and body movements, and then \textbf{U}nites those to learn a joint bi-directional distribution and directly predicts combined coefficients by a single output head. 
Our treatment on the output end enjoys three-fold benefits: 
1) By imposing supervision on the individual branches, it respects the distinctiveness between the two and ensures high-quality feature learning; 
2) The learned features enable bi-directional communication between branches in the latent space, fully exploiting the modeling capacity and facilitating the cost of association in the explicit space~\cite{chen2024diffsheg}; 
3) The final unification allows for a single-head generation, further enhancing the overall harmony on top of jointly learned features from each branch. 
Last but not least, combined with our design on the input end, DU-Trans also enables more harmonious customizable adaption.

To conclude, we have established \cb, a novel framework for \texttt{co}-speech holistic 3D human \texttt{m}otion generation and efficient customiza\texttt{b}le adaption, both in harm\texttt{o}ny. 
Interestingly, our framework echoes its aberration -- like a jazz band, it emphasizes harmony during training and inference (practice and play). 
Moreover, a well-trained jazz band can promptly incorporate with a new leader and/or play a new song with proper rehearsal. 
Similarly, $\cb$ can be adapted to new identities (leader) and/or emotions (a genre of songs) with efficient fine-tuning (rehearsing) as well. 
To validate the above, we comprehensively perform both quantitative and qualitative evaluations in BEAT2~\cite{liu2024emage} and SHOW~\cite{yi2023generating} datasets. 
Our proposed $\cb$ is highly effective in generating harmonious motions but also efficient in identity and emotion adaptation, \ie, our results significantly outperform others on the holistic metric FMD and also attain state-of-the-art performance in one-hour fine-tuning (about 5\% of the time required for training from scratch) with only updates about 10\% parameters.

In summary, our technical contributions are as follows:
\begin{itemize}
\item We reexamine the MIMO nature of co-speech holistic 3D human motion generation and focus on reducing its complexity for a harmonious and adaptive framework.

\item We propose DU-trans, which first captures the unique characteristics of the face and gesture for synchronization, then learns the joint distribution of them and directly predicts the combined coefficients for harmony.

\item We propose \texttt{X}-Adapter, which facilitates the fast and seamless adaptation of a pretrained neutral talking body to stylized versions or other different identities.

\item Extensive experiments on SHOW~\cite{yi2023generating} and BEAT2~\cite{liu2022beat} datasets confirm that our approach can realize the SOTA performance. Detailed
analysis validates the superiority of our method in motion quality and transfer efficiency.

\end{itemize}

\section{Related Work}

\subsection{Speech-Driven Body Motion Generation}
Holistic body motion generation from speech~\cite{chen2024diffsheg, liu2024emage, yi2023generating, mughal2024convofusion, liu2024towards, ng2024audio} encompasses the coordinated creation of movements for three key body parts: the face, hands, and body. However, most efforts only
consider parts of the human body rather than the holistic
body. For speech-driven facial movement generation, it is often referred to as talking face generation~\cite{shen2023difftalk, xu2023high, xu2023multimodal, peng2023emotalk, fan2022faceformer}, whether in 2D or 3D, is a vibrant field that involves creating animated faces that can mimic human speech and expressions. Recent 3D facial animation leverage blendshapes~\cite{peng2023emotalk, villanueva2022voice2face, zhang20213d} or 3D meshes~\cite{cudeiro2019capture, fan2022faceformer, peng2023selftalk} as the structure representation to control the lip shapes and capture speech nuances. Considering the complexity of mapping speech to facial expressions, probabilistic models such as VQ-VAE~\cite{van2017neural, ng2022learning, xing2023codetalker, yi2023generating} and diffusion models~\cite{ho2020denoising, aneja2024facetalk, chen2023diffusiontalker, stan2023facediffuser, sun2024diffposetalk, thambiraja20233diface} have been employed to predict the distribution of facial expressions derived from speech signals. Similarly, methods for speech-driven gesture generation~\cite{ahuja2020style, li2021audio2gestures, qi2024emotiongesture, yang2023diffusestylegesture, zhu2023taming, liu2022beat, ao2023gesturediffuclip, liu2022learning} are also designed to estimate the mapping or probability distribution of body and hand motion with the help of various condition modalities, including acoustic features, linguistic characteristics, speaker identities, and emotions. 

Recently, Habibie \emph{et al.}~\cite{habibie2021learning} first utilized a CNN-based framework to generate 3D facial meshes and 3D key points of the body and hands simultaneously. However, they overlook the coordination between body parts, and at the same time, deterministic models also lead to a lack of diversity. Thus, subsequent works all resort to a generative model to incorporate
diversity into motion generation. TalkSHOW~\cite{yi2023generating} is built upon the VQ-VAE and designs a cross-conditioned mechanism between the body and hand motions to keep the synchronization of the gesture, but they treat facial expression estimation as an independent task. To address the coordination issue between expression and gestures, DiffSHEG~\cite{chen2024diffsheg} and EMAGE~\cite{liu2024emage} divide two encoders, one for expression and one for gesture, and establish a path for unidirectional information flow between each. ProbTalk~\cite{liu2024towards} jointly models the holistic motion with the help of PQ-VAE~\cite{wu2019learning} in a unified manner. Nevertheless, current methods fail to learn harmonious relationships between various body parts while maintaining the unique distribution of each. In our work, we explore the feasibility of this by a divide and unite mechanism for highly synchronized and coordinated full-body motions. Moreover, existing approaches~\cite{chen2024diffsheg, liu2024emage, yi2023generating} typically require the definition of identities and emotions during training to guide the learning of specific characteristics. When the network encounters unseen identities and emotional styles, it may fail to generalize effectively, necessitating complete retraining or fine-tuning of the network with additional data to accommodate new conditions. In contrast, we introduce an efficient adaptation strategy that converts a pretrained model into various customized ones, enhancing the flexibility and applicability of our approach.

\subsection{Parameter-Efficient Fine-Tuning} 
Unlike conventional fine-tuning, which updates all parameters, Parameter-Efficient Fine-Tuning (PEFT) has shown impressive results across various tasks by updating only a subset of parameters, with the three primary methods being Adapter, Prefix-tuning, and LoRA. First, the concept of adapters is first introduced by the Serial Adapter~\cite{houlsby2019parameter} model. This model enhances each Transformer block by incorporating two additional adapter modules, one after the self-attention layer and the other after the feed-forward neural network (FFN) layer. Parallel Adapter~\cite{he2021towards}, on the other hand, restructures these sequential layers into a parallel side network that operates in tandem with each Transformer sublayer. Other works, such as CIAT~\cite{zhu2021counter} and CoDA~\cite{lei2023conditional} are also adopts a parallel adapter. Second, Prefix-tuning~\cite{li2021prefix} introduces tunable vectors that are prepended to keys and values across all attention layers. Concurrent work p-tuning~\cite{liu2021p} and prompt-tuning~\cite{lester2021power} apply learnable vectors only at the initial word embedding layer rather than all layers to enhance training and inference efficiency. Although these concepts can be prone to slow convergence, they have been employed for various downstream tasks~\cite{choi2023codeprompt, wu2022adversarial}. Third, LoRA~\cite{hu2022lora} adopts a low-rank approximation to update weight matrices in the attention. Several subsequent studies~\cite{liu2024dora, zhang2023adalora, valipour2022dylora} aim to improve LoRA's performance in various aspects. Our \texttt{X}-Adapter is largely based on the first choice and follows the parallel design. Notably, EAT~\cite{gan2023efficient} employs a pre-train-and-fine-tune approach similar to ours, but they extend a trainable encoder branch for fine-tuning directly, which is neither as efficient nor as effective as PEFT. 

In this work, we integrate PEFT into the domain of holistic 3D human motion generation for fast and flexible adaptation, which is the first attempt in this field. From an application perspective, holistic 3D human motion finds applications across various scenarios. However, currently there is no single universal model capable of handling all demands. A fallback approach is to offer only some customized services, which, however, lacks flexibility and limits the scope of application. Our motivation is to have a powerful basic model that can cover most demands, and combine it with PEFT for fast fine-tuning. This allows the basic model to quickly adapt to various fine-tuning data within a short period, thereby covering a wide range of scenarios. We believe that the combination of pretraining and fine-tuning is a viable approach to advancing the practical application of motion generation. More methodological and experimental details are presented in the Secs.~\ref{sec:3}-\ref{sec:4}.

\section{Method}
\label{sec:3}
\begin{figure*}[t!]
	\centering
	\includegraphics[width=0.85\textwidth]{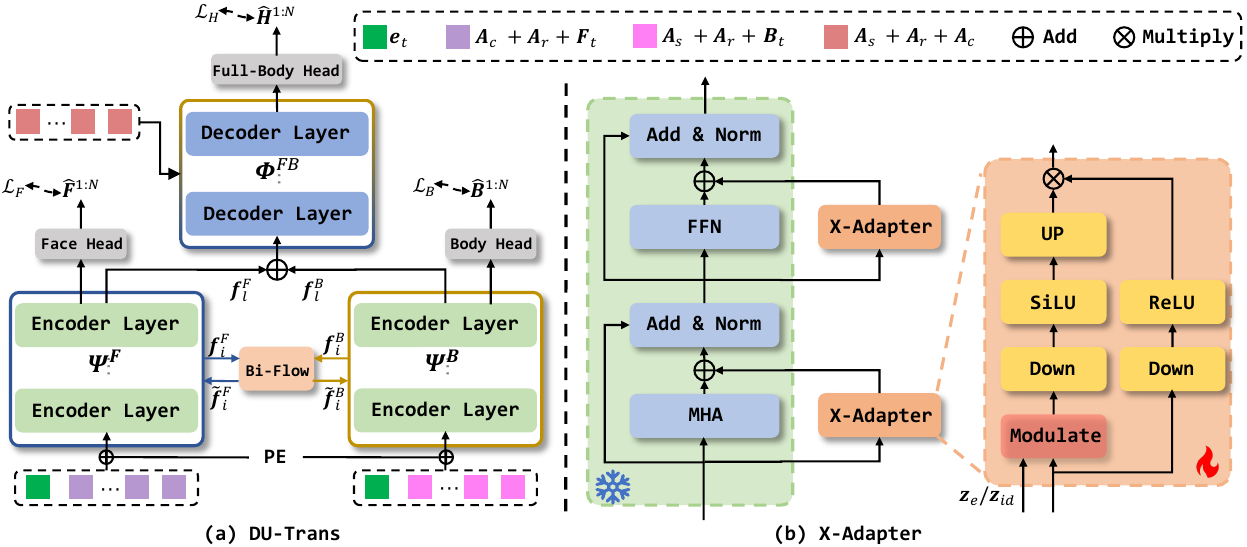}
	\caption{\textbf{Architecture overview of \cb.} The basic architecture named DU-Trans (a) first introduces two transformer encoders $\boldsymbol{\Psi}^F$, $\boldsymbol{\Psi}^B$ incorporated with auxiliary losses $\mathcal{L}_F$, $\mathcal{L}_B$ and Bi-Flow to help model their respective distributions, obtaining two sets of discriminative features $\boldsymbol{f}^F_l$, $\boldsymbol{f}^B_l$. Subsequently, it merges these two features and inputs them into the decoder $\boldsymbol{\Phi}^{FB}$ to learn the joint distribution, and directly uses a single head to predict synchronized and coordinated face and body coefficients. Then, \texttt{X}-Adapter (b) is the central module for achieving identity customization and emotion transfer, and it is simply inserted in parallel into the MHA and FFN layers of the two encoders. Note that this adapter is a general structure suitable for both identity and emotion, offering better controllable generation through conditions $\boldsymbol{z}_e$ and $\boldsymbol{z}_{id}$.}
	\label{fig:pipeline}
\end{figure*}
Our proposed $\cb$ framework aims to deliver harmonious holistic motion generation as well as efficient customizable adaption with respect to new identity and/or emotion. 
Before diving into technical details, we first give a preliminary overview on holistic human motion generation in Section~\ref{sec:pre}. 
Then we introduce DU-Trans in Section~\ref{Sec:3.2}, which is key to creating synchronized and coordinated full-body talking motions. 
After that, we introduce \texttt{X}-Adapter for efficient emotion style transfer and rapid personalization in Section~\ref{Sec:3.3}. 
Finally, we describe how \texttt{X}-Adapter can be used for conditional editing in Section~\ref{sec:edit}

\subsection{Preliminary}\label{sec:pre}
\noindent\textbf{Holistic Human Shape Model.} Following~\cite{yi2023generating,chen2024diffsheg,liu2024emage}, we use  SMPLX~\cite{pavlakos2019expressive} as generation representation, which is a well-known parametric model for characterizing holistic expressive human shapes, including detailed facial expression, hand gesture and body pose. 
This model is widely utilized in computer graphics and virtual reality, providing a high degree of realism and flexibility for character creation and motion animation.
Starting with a template mesh consisting of $n$ vertices and a fixed triangulation, SMPLX associates parameters and human shapes via function
$M(\boldsymbol{\rho},\boldsymbol{\omega},\boldsymbol{\psi}): \mathbb{R}^{| \boldsymbol{\rho}| + |\boldsymbol{\omega}| + |\boldsymbol{\psi} |} \rightarrow \mathbb{R}^{3n}$. 
Namely, given a set of parameters $(\boldsymbol{\rho},\boldsymbol{\omega},\boldsymbol{\psi})$, the function returns coordinates of a human shape in a fixed order, which, together with the pre-fixed triangulation, forms a plausible human mesh. 
The parameters carry different semantics: identity $\boldsymbol{\rho} \in \mathbb{R}^{300}$ captures the specific shape characteristics of the body; pose $\boldsymbol{\omega} \in \mathbb{R}^{J \times 3}$ encodes rotations around a defined set of joints ($J = 55$), which allows details as intricate as finger articulation; facial expression $\boldsymbol{\psi} \in \mathbb{R}^{100}$ represents a wide array of facial movements for expressive animations.  
Thanks to SMPLX, we can cast holistic human motion generation as the respective parameter generation. 

\noindent\textbf{Diffusion Models.}
Our approach employs diffusion models, encompassing diffusion and denoising stages. 
With a given distribution of motion clips, our objective is to train a model with parameters $\theta$ to approximate the initial state $\boldsymbol{x}_0^{1:N}$ (noise-free sequence of $\boldsymbol{x}^{1:N}$, $N$ is the sequence length). 
During the diffusion phase, the model incrementally degrades the input data $\boldsymbol{x}_0^{1:N} \sim p(\boldsymbol{x}_0^{1:N})$ following a set schedule $\beta_t \in (0,1)$, culminating in an isotropic Gaussian distribution across $T$ steps. 
Each step of the forward transition can be represented as:
\begin{equation}
\begin{aligned}
    q\left(\mathbf{x}_t^{1:N} \mid \mathbf{x}_{t-1}^{1:N}\right)=\mathcal{N}\left(\mathbf{x}_t^{1:N} ; \sqrt{1-\beta_t} \mathbf{x}_{t-1}^{1:N}, \beta_t \mathbf{I}\right).
\end{aligned}
\end{equation}
Conversely, in the denoising phase, the model is trained to reverse the noising process, thereby converting noise back into the actual data distribution during inference. 
The backward transition is:
\begin{equation}
\begin{aligned}
p_\theta\left(\mathbf{x}_{t-1}^{1:N} \mid \mathbf{x}_t^{1:N}\right)=\mathcal{N}\left(\mathbf{x}_{t-1}^{1:N} ; \mu_\theta\left(\mathbf{x}_t^{1:N}, t\right), \Sigma_\theta\left(\mathbf{x}_t^{1:N}, t\right)\right).
\end{aligned}
\end{equation}
We follow Ho \emph{et al.}~\cite{ho2020denoising} to model the mean $\mu_\theta\left(\mathbf{x}_t^{1:N}, t\right)$ of the reverse distribution while keeping the variance $\Sigma_\theta\left(\mathbf{x}_t^{1:N}, t\right)$ fixed. 
Instead of predicting the noise $\boldsymbol{\epsilon}_t$, we follow Ramesh \emph{et al.}~\cite{ramesh2022hierarchical} and predict the signal $\hat{\boldsymbol{x}}_0^{1:N}$ itself with the simple objective~\cite{ho2020denoising}:
\begin{equation}
\begin{aligned}
\mathcal{L}_{\text{simple }}=\mathbb{E}_{t, \boldsymbol{x}_0}\left[\boldsymbol{x}_0^{1:N}-\hat{\boldsymbol{x}}_0^{1:N}\right].
\end{aligned}
\label{eq:3}
\end{equation}

\noindent\textbf{Adapters}~\cite{houlsby2019parameter} are one of the state-of-the-art techniques in Parameter-Efficient Fine-Tuning (PEFT), implemented by embedding compact, auxiliary layers into the Transformers. 
The adapter layer typically compresses the input $\boldsymbol{h}$ into a lower-dimensional space via a down-projection with matrix $\boldsymbol{W}_{\text{down}} \in \mathbb{R}^{d \times r}$, constrained by the bottleneck dimension $r$ to minimize parameter count. 
It then applies a non-linear activation function $\delta(\cdot)$, before expanding the features back with an up-projection using $\boldsymbol{W}_{\text{up}} \in \mathbb{R}^{r \times d}$. 
This bottleneck module is connected to the
original pre-trained models through the residual connection. 

\subsection{DU-Trans}\label{Sec:3.2}
Fig.~\ref{fig:pipeline}(a) shows an illustration of DU-Trans. 
In the following, we begin by describing the extraction of comprehensive audio features. 
Then we present details of the network architecture. 
Finally, we introduce the training loss terms.

\noindent\textbf{Audio Feature Extraction.} 
In this part, we follow the common practice~\cite{chang2023hierarchical, sun2025beyond} to decompose input speech signal into content, rhythm, and semantics. 
As suggested by recent progress~\cite{ng2024audio}, the local lip motion is strongly correlated with the input audio content, and the global facial expression is related to the audio rhythm. 
While the body has a weaker correlation with the content, yet is intricately connected to audio semantics and rhythm~\cite{ao2022rhythmic}.
Early approach~\cite{yi2023generating} leverages MFCC~\cite{muda2010voice} to encode speech, which falls short of capturing rich speech information and struggles with disentangling each component.
Therefore, we resort to recent advances in audio pre-trained models.  
Specifically, we use wav2vec 2.0~\cite{baevski2020wav2vec} trained on Automatic Speech Recognition (ASR) task as our audio content extractor, obtaining $\boldsymbol{A}_c^{1:N} \in \mathbb{R}^{1024 \times N}$, which primarily retains the phonemes and filters out irrelevant information. 
Regarding rhythm, we choose the JDC network~\cite{kum2019joint} trained on LibriSpeech~\cite{panayotov2015librispeech} to predict acoustic rhythm $\boldsymbol{A}_r^{1:N} \in \mathbb{R}^{1 \times N}$. 
For semantics, we first follow~\cite{ao2022rhythmic} to align words with the corresponding speech and convert the text into frame-level features (SHOW~\cite{yi2023generating} dataset requires this processing while BEAT2~\cite{liu2022beat} does not). 
Then we take the aligned text as input and use the pretrained model of BERT~\cite{devlin2019bert} to encode $\boldsymbol{A}_s^{1:N} \in \mathbb{R}^{1536 \times N}$.

\noindent\textbf{Architecture.} 
Given the markedly different audio-related dynamics of facial expressions and gestures, it is essential to independently model these components for enhanced synchronization. 
However, this approach alone risks neglecting the intrinsic connections between them and could cause disjointed coordination. 
Previous works~\cite{chen2024diffsheg, liu2024emage} have recognized this issue, but they only introduce uni-flow and use multiple heads to predict individual coefficients still hindering overall coordination. 
In contrast, we propose a divide-and-unite strategy that respects the distinct modeling needs of each component while implicitly accounting for their interrelationships and directly predicting a unified set of coefficients, thus possessing synchronization and coordination in a unified framework. 
Specifically, we combine this strategy with a diffusion model. 
As shown in Fig.~\ref{fig:pipeline}(a), we first implement four key designs to learn the distinct dynamic characteristics of the face and body: 

\noindent 1) Two separate Transformer encoders $\boldsymbol{\Psi}^{F}$, $\boldsymbol{\Psi}^{B}$ are employed to respectively model the features of face and body, allowing for specialized feature modeling for face and body movements; 

\noindent 2) Each branch is designed to receive features with which it has a strong correlation. 
The face branch takes as input audio content $\boldsymbol{A}_c^{1:N}$, rhythm $\boldsymbol{A}_r^{1:N}$, and noise face sequences $\boldsymbol{F}_t^{1:N}$, with time-step information $\boldsymbol{e}_t$ also integrated into the input sequence. 
Formally:
\begin{equation}
\begin{aligned}
\boldsymbol{f}_l^F = \boldsymbol{\Psi}^{F}(\text{PE}+[\boldsymbol{e}_t, \boldsymbol{A}_c^{1:N} + \boldsymbol{A}_r^{1:N} + \boldsymbol{F}_t^{1:N}]),
\end{aligned}
\end{equation}
where $[\cdot]$ means concatenation, $l$ is the number of encoder layers, and PE is the positional embedding. 
The body branch, on the other hand, receives semantic information rather than phonetic content:
\begin{equation}
\begin{aligned}
\boldsymbol{f}_l^B = \boldsymbol{\Psi}^{B}(\text{PE}+[\boldsymbol{e}_t, \boldsymbol{A}_s^{1:N} + \boldsymbol{A}_r^{1:N} + \boldsymbol{B}_t^{1:N}]).
\end{aligned}
\end{equation}

\noindent 3) A Bi-Flow layer built upon the cross-attention mechanism~\cite{vaswani2017attention} is introduced to preliminarily model the relationship between the two branches, capturing holistic dynamic priors to enhance the performance of each component. 
We take face to body data-flow for example, the query $\boldsymbol{Q}_B$ is extracted by linear projection from $\boldsymbol{f}^B_i$ ($i$ is the layer index), and the key and value $\boldsymbol{K}_F$, $\boldsymbol{V}_F$ is extracted from $\boldsymbol{f}^F_i$ in the same way. 
To obtain the updated body feature $\tilde{\boldsymbol{f}}^B_i$,
\begin{eqnarray}
    &\boldsymbol{f}^{F \rightarrow B}_i = \text{softmax}(\frac{\boldsymbol{Q}_B(\boldsymbol{K}_F)^T}{\sqrt{d}} )\boldsymbol{V}_F, \\ 
    &\tilde{\boldsymbol{f}}_i^B=\text{MLP}(\text{LN}(\boldsymbol{f}^{F \rightarrow B}_i))+\boldsymbol{f}_i^B,
\end{eqnarray}
where $\text{MLP}$ and $\text{LN}$ is a MLP block and a LayerNorm, $\sqrt{d}$ is a scaling factor. 

\noindent 4) Our architecture consists of three output heads: Two output heads are used to predict independent coefficients for face and body under auxiliary supervisions (\emph{c.f.} Loss Functions paragraph below); One output head is for holistic generation.
By summing the face features $\boldsymbol{f}_l^F$ and body features $\boldsymbol{f}_l^B$ output by the two encoders and feeding the sum into the single decoder $\boldsymbol{\Phi}^{FB}$ to implicitly model their interrelations. 
During training, all output heads are optimized with regarding training loss, while during inference only the last head is activated to deliver coordinated holistic human motions.
Formally,
\begin{equation}
\begin{aligned}
\boldsymbol{f}^{FB} = \boldsymbol{\Phi}^{FB}(\boldsymbol{f}_l^{F}+\boldsymbol{f}_l^{B}, \boldsymbol{A}_s^{1:N} + \boldsymbol{A}_r^{1:N} + \boldsymbol{A}_c^{1:N}).
\end{aligned}
\end{equation}

\noindent\textbf{Loss Functions.}
During the training phase, our model outputs three predicted parameters: facial expression $\hat{\boldsymbol{F}}^{1:N} \in \mathbb{R}^{N \times 100}$ and body gestures $\hat{\boldsymbol{B}}^{1:N} \in \mathbb{R}^{N \times 165}$ output from two encoder, and the combined one $\hat{\boldsymbol{H}}^{1:N} \in \mathbb{R}^{N \times 265}$ output from a single decoder. Each is supervised by two loss functions, simple loss $\mathcal{L}_{\text{simple }}$ following the Eq.~\ref{eq:3} and velocity loss $\mathcal{L}_{\text{vel}}$. Formally, we take the holistic body as an example:
\begin{eqnarray}
&\mathcal{L}_{\text{vel}}=\frac{1}{N-1} \sum_{i=1}^{N-1}\left\|\left(\boldsymbol{H}^{i+1}-\boldsymbol{H}^i\right)-\left(\hat{\boldsymbol{H}}^{i+1}-\hat{\boldsymbol{H}}^i\right)\right\|_2^2,\\
&\mathcal{L}_{\text{simple }}=\mathbb{E}_{t, \boldsymbol{H}}\left[\boldsymbol{H}^{1:N}-\hat{\boldsymbol{H}}^{1:N}\right],\\
&\mathcal{L}_H=\mathcal{L}_{\text{vel}}+\mathcal{L}_{\text{simple}}.
\end{eqnarray}
Overall, our training loss is:
\begin{equation}
\begin{aligned}
\mathcal{L} = \mathcal{L}_H + \lambda_F \mathcal{L}_F + \lambda_B \mathcal{L}_B,
\end{aligned}
\end{equation}
where $\lambda_F$ and $\lambda_B$ are set to 0.5 and 0.5, respectively. $\mathcal{L}_F$ and $\mathcal{L}_B$ serve as two auxiliary losses during training.

\subsection{\texttt{X}-Adapter}
\label{Sec:3.3}
In this section, we present \texttt{X}-Adapter, as shown in Fig.~\ref{fig:pipeline}(b) and Alg.~\ref{algo:algo}, for efficient fine-tuning.

\noindent\textbf{Architecture.} Based on the vanilla adapter~\cite{houlsby2019parameter} described in Sec.~\ref{sec:pre}, we make some modifications and propose the \texttt{X}-Adapter.

\noindent 1) To balance the task-agnostic features generated by the original frozen branch and the task-specific features generated by the tunable bottleneck branch, we do not rely on a simple scalar hyperparameter as a scale. 
Instead, we adopt a parallel down-projection layer with matrix $\boldsymbol{W}_{\text{s}} \in \mathbb{R}^{d \times 1}$ to dynamically generate a scale factor, named Dy-Scale $\boldsymbol{s}_d \in \mathbb{R}^{N}$ based on the input motion sequences. 
Importantly, we follow this with a ReLU activation to select the positive scale and set the rest to zero. 
Because only significant local motion tokens require adjustment during fine-tuning, it should depend on the unique characters of each input feature:
\begin{equation}
\begin{aligned}
\boldsymbol{s}_d=\text{ReLU}(\boldsymbol{W}_{s}\boldsymbol{h}).
\end{aligned}
\end{equation}
2) We insert a modulation layer $\boldsymbol{\mathcal{M}}$ before the down-projection phase to seamlessly infuse task-specific conditions $\texttt{X}$ ($\boldsymbol{z}_s$ or $\boldsymbol{z}_{id}$) into the adapter module. Striking a balance between performance and parameter efficiency, this modulation is elegantly achieved through the use of addition alone:
\begin{equation}
\begin{aligned}
\boldsymbol{\mathcal{M}}(\boldsymbol{h})=\texttt{x} + \boldsymbol{h}.
\end{aligned}
\end{equation}
3) These adapters are inserted at the multi-head attention (MHA) and feed-forward network (FFN) in a parallel manner, which preserves original features via a separate branch while aggregating updated context through element-wise scaling. Overall,
\begin{equation}
\begin{aligned}
\text{\texttt{X}-Adapter}(\boldsymbol{h})=\boldsymbol{s}_d \times \boldsymbol{W}_{\text{up}}\times \delta\left(\boldsymbol{W}_{\text{down}} \boldsymbol{\mathcal{M}}(\boldsymbol{h})\right)+\boldsymbol{h},
\end{aligned}
\end{equation}
where $\delta(\cdot)$ is a SiLU activation in our model. In the fine-tuning process, we selectively update only the newly introduced parameters (orange blocks in Fig.~\ref{fig:pipeline}(b)), leaving the rest (blue blocks) unchanged.

\begin{algorithm}[t]
    \caption{PyTorch-like code of \texttt{X}-Adapter.}
    \label{algo:algo}
    \footnotesize
\begin{alltt}
import torch.nn as nn
class XAdapter(nn.Module):
    def __init__(self, rank, d_model):
        super().__init__()
        self.down_proj = nn.Linear(d_model, rank)
        self.non_linear_func = nn.SiLU()
        self.up_proj = nn.Linear(rank, d_model)

        self.dy_scale = nn.Linear(d_model, 1)
        self.relu = nn.ReLU()
        
    def forward(self, x, cond):
        \color{ForestGreen}# Dy-Scale\color{black}
        self.scale = self.relu(self.dy_scale(x))
        \color{ForestGreen}# Modulate\color{black}
        x = x + cond
        \color{ForestGreen}# Common Adapter Processing\color{black}
        down = self.down_proj(x)
        down = self.non_linear_func(down)
        up = self.up_proj(down)
        \color{ForestGreen}# Update\color{black}
        output = up * self.scale
        return output
\end{alltt}
\end{algorithm}

\subsection{Conditional Editing based on \texttt{X}-Adapter}\label{sec:edit}
Current holistic methods~\cite{liu2024emage, chen2024diffsheg} train on all emotional data under a fixed identity to achieve diverse outputs, but this approach sacrifices the capability for editing. In conjunction with identity- and emotion-agnostic audio processing in Sec.~\ref{Sec:3.2}, the \texttt{X}-Adapter supplements extra conditions during training, thus facilitating further editing during inference.
In the following, we describe how to embed emotion and identity information from external sources into latent codes, obtaining $\boldsymbol{z}_e$ and $\boldsymbol{z}_{id}$.

\begin{figure}[t!]
	\centering
	\includegraphics[width=0.48\textwidth]{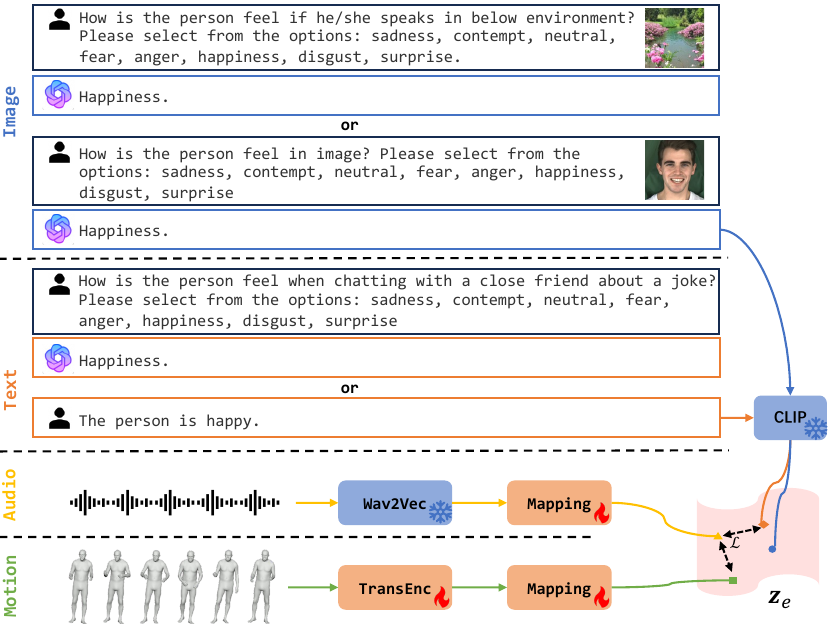}
	\caption{\textbf{Overview of multi-modal emotion space within the CLIP domain.} In this space, text and image benefit from train-free GPT and CLIP, while audio and motion undergo training to align latent representations with corresponding emotional text embeddings in CLIP space. That is, the emotion cues extracted from a happy audio and happy motion sequence are supervised by the embedding of emotional text prompt \texttt{The person is happy}.}
	\label{fig:multimodal}
\end{figure}

\noindent\textbf{Emotion Condition.}
Inspired by GestureDiffuCLIP~\cite{ao2023gesturediffuclip}, we aim to align emotion cues to the CLIP domain~\cite{radford2021learning}, and enable the use of various multi-modal prompts to indicate emotions. As a result, any accessible modality can serve as guidance, making the application more flexible. Additionally, it can support unknown emotion guidance, thanks to the rich semantics provided by CLIP. As shown in Fig.~\ref{fig:multimodal}, our system allows users to describe the desired emotion by text, image, audio, and motion sequence. Specifically, for the text prompt, we can directly specify emotional conditions, \eg, \texttt{The person is happy} or provide descriptive prompts for GPT to generate the corresponding emotional text prompts, \eg, \texttt{How is the person feel when chatting with a close friend about a joke}, and then extract corresponding embeddings by CLIP encoder. Similarly, we align image prompts with the emotional text prompts by GPT. For example, \texttt{How is the person feel [a picture of a happy face]} or \texttt{How is the person feel if he/she speaks in [a picture of a beautiful garden]}. The above two are train-free. When querying, we instruct GPT to select the most fitting emotional category from the following options \texttt{[sadness, contempt, neutral, fear, anger, happiness, disgust, surprise]}. For the last two, we employ a pretrained wav2vec2.0~\cite{baevski2020wav2vec} model, specialized in emotion recognition, to distill effective information from audio. In parallel, we adopt the framework of Motionclip~\cite{tevet2022motionclip} to learn emotional features from motion sequences. These output cues are further mapped into the CLIP domain and constrained by computing cosine similarity with the CLIP embeddings of emotional text prompts. The above two are trained in BEAT2~\cite{liu2022beat} dataset. Consequently, our training uses only the embedding derived from text, yet inference allows various modalities as guidance.

\noindent\textbf{Identity Condition.}
Identity is a crucial factor that affects long-term face and body movements. For example, in BEAT2 dataset~\cite{liu2022beat}, Scott is typically quite excited and speaks with broad facial and body gestures, while Nidal tends to be solemn, using only subtle hand movements when speaking. According to the work~\cite{wu2021imitating}, the identity is closely related to the variance of facial and gesture movements inside each video, we can calculate standard derivation $\sigma(\cdot)$ with respect to frame $k$ of ($\boldsymbol{F}(k)$, $\boldsymbol{B}(k)$, $\frac{\partial\boldsymbol{F}(k)}{\partial k})$, $\frac{\partial\boldsymbol{B}(k)}{\partial k})$. Formally, given arbitrary video with the reconstructed parameters series $\boldsymbol{F}(k)$, $\boldsymbol{B}(k)$, the identity codes $\boldsymbol{z}_{id}^F$ and $\boldsymbol{z}_{id}^B$ are defined as:
\begin{eqnarray}
    &\boldsymbol{z}_{id}^F = \text{MLP}([\sigma(\boldsymbol{F}(k)), \sigma(\frac{\partial\boldsymbol{F}(k)}{\partial k}))]),\\
    &\boldsymbol{z}_{id}^B = \text{MLP}([\sigma(\boldsymbol{B}(k)), \sigma(\frac{\partial\boldsymbol{B}(k)}{\partial k}))]),
\end{eqnarray}
where MLP performs dimensional mapping, which is trained along with adapters. Given some video data of a particular identity, we select a clip to calculate the style codes, which are shared for all inputs across subsequent training and testing.

\section{Experiments}
\label{sec:4}
\subsection{Experimental Setup}

\subsubsection{Datasets}

\noindent\textbf{BEAT2} is proposed in EMAGE~\cite{liu2024emage}, which is built on the original BEAT dataset~\cite{liu2022beat} (containing 76 hours of data for 30 speakers). 
In particular, BEAT2 transfers the complex annotation of the latter to the standard mesh representation, along with paired audio and text transcripts. We employ the BEAT2-standard portion and adopt $85\%, 7.5\%$, and $7.5\%$ for each identity. 
In the following experiments, we utilize four identities, Speaker-1 (Wayne), Speaker-2 (Scott), Speaker-11 (Nidal), and Speaker-23 (Hailing).

\noindent\textbf{SHOW} ~\cite{yi2023generating} is a high-quality audio-visual dataset, which consists of $26.9$ hours of in-the-wild talkshow videos from 4 speakers with 3D body meshes at 30fps, and their synchronized audio at a 22K sample rate. 
We select video sequences longer than 10 seconds and divide the dataset into 80\%, 10\%, and 10\% as train, validation, and test splits.

\subsubsection{Implementation Details}
We implement our network using PyTorch. The basic DU-Trans contains seven encoder layers and one decoder layer. The hidden dimension of all
transformer layers is 512. The bottleneck dimension of \texttt{X}-Adapter is set to 128. Our diffusion model employs a cosine noise schedule, with diffusion steps set to 1000 for training and inference. During pretraining, the learning rate is set to 1e-4 using the ADAM optimizer with $[\beta_1, \beta_2]=[0.9, 0.99]$ and adjusted to 1e-3 for fine-tuning experiments. The batch size is consistently set to 128. We train our pretraining model on a single NVIDIA A100 GPU for one day, completing 100,000 iterations. For fine-tuning, we perform 5,000 iterations within one hour. The length of the input motion sequences is
600.

\begin{table}[t!]
   \caption{\textbf{Quantitative comparison with SOTA methods on BEAT2 dataset.} The "$\downarrow$" means the lower, the better, and vice versa. \textbf{Bold} and \underline{underline} represent optimal and suboptimal results. The * indicates training from scratch (pretraining), while the $\dagger$ signifies fine-tuning the emotional model from the neutral pre-trained one. For simplicity, we report MSE$\times10^{-8}$ and LVD$\times10^{-5}$ as EMAGE.}
  \centering
  \scriptsize
   \renewcommand\arraystretch{1.2}
   \setlength\tabcolsep{6pt}
   \begin{tabular}{C{2pt}C{1pt}C{52pt}C{15pt}C{15pt}C{15pt}C{15pt}C{15pt}C{15pt}}
      \toprule
       Dataset & &Method  & FMD$\downarrow$  & FGD$\downarrow$    & BC$\uparrow$  & DIV$\uparrow$  & MSE$\downarrow$ & LVD$\downarrow$ \\
      \midrule
      \multirow{19}{*}{{\rotatebox[origin=c]{90}{BEAT2 (Scott)}}}
      &\multirow{9}{*}{{\rotatebox[origin=c]{90}{Neutral}}}
      & FaceFormer~\cite{fan2022faceformer} & - & - & - & - &7.725 & 7.619\\ &&CodeTalker~\cite{xing2023codetalker} & - & - & - & - &8.133 & 7.764\\
      &&CaMN~\cite{liu2022beat} & 1.546 & 0.668 & 0.6712 & 10.36 & - & -\\ &&DSG~\cite{yang2023diffusestylegesture} & 1.677 & 0.891 & 0.7396 & \underline{10.93} & - & -\\
      &&Habibie \emph{et al.}~\cite{ahuja2020style} & 1.896 & 0.902 & 0.7842 & 8.669 & 8.658 & 8.102 \\ &&TalkSHOW~\cite{yi2023generating} & 1.321 & 0.871 & 0.7776 & 10.42 & \underline{7.476} & 7.765\\
      & & EMAGE~\cite{liu2024emage} & \underline{1.287} & \underline{0.662} & \underline{0.7907} & \textbf{13.01} & 7.703 & \underline{7.460}\\
      \cmidrule(lr){3-9}
      & & Ours* & \textbf{1.098} & \textbf{0.563} & \textbf{0.8023} & 10.48  & \textbf{5.098} & \textbf{6.005}  \\
      \cmidrule(lr){2-9}
      &\multirow{10}{*}{{\rotatebox[origin=c]{90}{Emotional}}}
      & FaceFormer~\cite{fan2022faceformer} & -  & - & - & - & 7.814	& 7.657\\
      & & CodeTalker~\cite{xing2023codetalker} & -  & - & - & - & 8.001 & 7.830\\
      & & CaMN~\cite{liu2022beat} & 1.594 & 0.657 & 0.6812 & 10.90 & - & -\\
      & & DSG~\cite{yang2023diffusestylegesture} & 1.640 & 0.885 & 0.7405 & \underline{11.05} & - & -\\
      & & Habibie \emph{et al.}~\cite{ahuja2020style} & 1.903 & 0.910 & 0.7797 & 8.761 & 8.698 & 8.143\\
      & & TalkSHOW~\cite{yi2023generating} & 1.310 & 0.858 & 0.7622 & 10.47 & \underline{7.531} & 7.612\\
      & & EMAGE~\cite{liu2024emage} & \underline{1.239} & \underline{0.656} & \underline{0.7990} & \textbf{13.09} & 7.723 & \underline{7.471} \\
      \cmidrule(lr){3-9}
      & & Ours$\dagger$ & \textbf{1.128} & \textbf{0.568} & \textbf{0.8003} & 10.63 & \textbf{5.015} & \textbf{6.408} \\
      \midrule
      \multirow{8}{*}{{\rotatebox[origin=c]{90}{BEAT2 (Nidal)}}}
      &\multirow{4}{*}{{\rotatebox[origin=c]{90}{Neutral}}}
      &TalkSHOW~\cite{yi2023generating} & 0.706 & 0.256 & 0.6213 & 5.693 & \underline{1.825} & 4.118\\
      & & EMAGE~\cite{liu2024emage} & \underline{0.688} & \underline{0.235} & \underline{0.6272} & \underline{6.057} & 1.983 & \underline{4.021}\\
      \cmidrule(lr){3-9}
      & & Ours* & \textbf{0.651} & \textbf{0.228} & \textbf{0.6391} & \textbf{6.062} & \textbf{1.685} & \textbf{3.768}  \\
      \cmidrule(lr){2-9}
      &\multirow{4}{*}{{\rotatebox[origin=c]{90}{Emotional}}}
      & TalkSHOW~\cite{yi2023generating} & 0.817 & 0.322 & 0.6284 & 6.796 & 2.897 & 4.821\\
      & & EMAGE~\cite{liu2024emage} & \underline{0.801} & \underline{0.314} & \underline{0.6405} & \textbf{7.308} & \underline{2.896} & \underline{4.705} \\
      \cmidrule(lr){3-9}
      & & Ours$\dagger$ & \textbf{0.783} & \textbf{0.300} & \textbf{0.6536} & \underline{7.125} & \textbf{2.664} & \textbf{4.442} \\
      \bottomrule
   \end{tabular}
   \label{tab:sota}
\end{table}

\subsubsection{Metrics}
We assess the quality of the generated motion from three aspects. For the whole body, we utilize \textbf{FMD}~\cite{chen2024diffsheg} to measure the difference between the distributions of the generated holistic motion and ground truth in feature space. It can indicate the overall quality of the holistic motions and the coordination among different body parts. For body gestures, we use \textbf{FGD}~\cite{yoon2020speech} to measure the distribution difference between generated gesture and ground truth.
For the above two metrics, we use a Skeleton CNN based encoder and a Full CNN-based decoder as the autoencoder pretrained network following~\cite{liu2024emage}. The network is pretrained on both BEAT2 and SHOW datasets.
\textbf{BC}~\cite{li2021ai} quantifies the alignment between the rhythm of the generated gesture and the beat of the audio. \textbf{DIV}~\cite{li2021audio2gestures} is a metric for measuring the variations of the synthesized gesture. For face expression, we employ two reconstruction metrics. The vertex \textbf{MSE}~\cite{xing2023codetalker} is calculated to determine the positional distance, and the vertex L1 difference \textbf{LVD}~\cite{yi2023generating} is used to assess the discrepancy between the ground truth and the generated facial vertices. Besides, since this task lacks clear ground, human objective evaluation is another main criterion in our method. We conduct an extensive user study to rate the generated motions by different methods in terms of coordination, coherence, and synchronization.

\begin{figure*}[t!]
	\centering
	\includegraphics[width=0.95\textwidth]{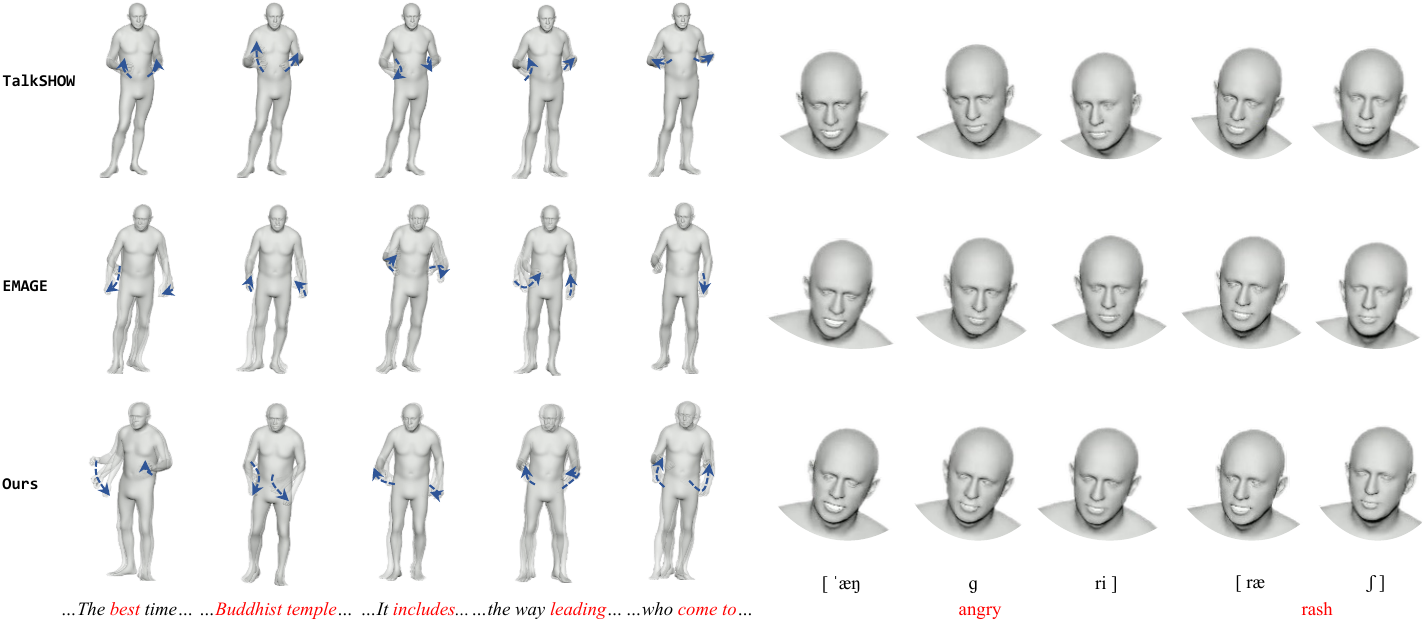}
	\caption{\textbf{Qualitative comparison with TalkSHOW and EMAGE on BEAT2 dataset.} The left part shows the holistic motions while the right presents a close-up of the expressions. Our method can generate expressions and gestures that are synchronized with the audio, particularly producing accurate and diverse gestures for rhythm, semantics, and specific concepts.}
	\label{fig:sota_beat}
\end{figure*}

\begin{figure*}[t!]
	\centering
	\includegraphics[width=0.95\textwidth]{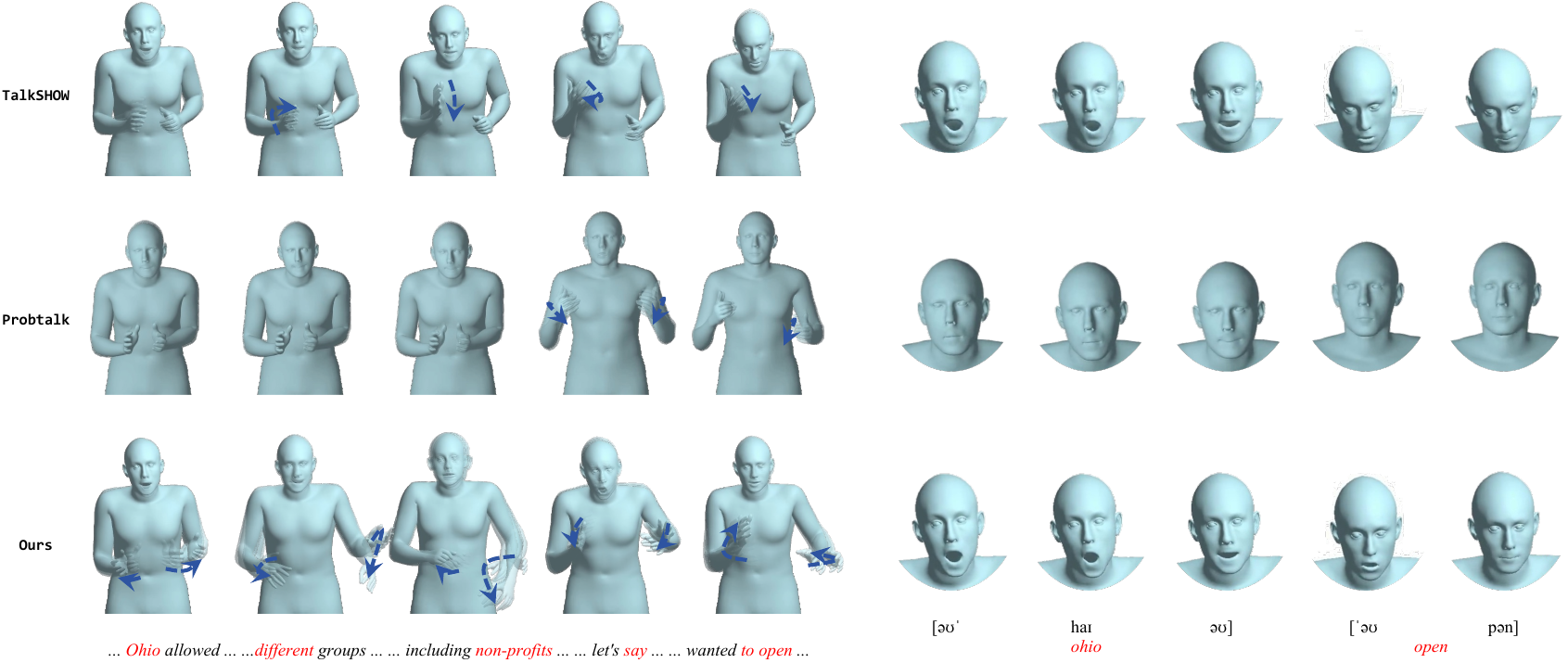}
	\caption{\textbf{Qualitative comparison with TalkSHOW and ProbTalk on SHOW dataset.} We utilize official checkpoints of competing models and follow the official rendering format for fair comparisons.}
	\label{fig:sota_show}
\end{figure*}

\subsubsection{Baselines}
For BEAT2 dataset~\cite{liu2024emage}, we compare our method with representative SOTA approaches in talking head generation: FaceFormer~\cite{fan2022faceformer} and CodeTalker~\cite{xing2023codetalker}, both of which are tailored for speech-driven 3D facial animation. Additionally, we evaluate against body gesture generation methods, CaMN~\cite{liu2022beat} which introduce the BEAT dataset constructed by using a commercial motion capture system, and DSG~\cite{yang2023diffusestylegesture}, a novel diffusion-based co-speech generation method. We reproduce their results in the standardized format of BEAT2 for the face and body respectively. Furthermore, we incorporate assessments of three recent holistic pipelines: the Habibie \emph{et al.}~\cite{ahuja2020style} and TalkSHOW~\cite{yi2023generating} are retrained in standardized format either, and EMAGE~\cite{liu2024emage} which directly uses their officially released weights for evaluation. On the SHOW dataset~\cite{yi2023generating}, we replace EMAGE with the recent ProbTalk~\cite{liu2024towards} because it offers an official checkpoint. Additionally, TalkSHOW also utilizes the officially released weights. The motion rendering follows the official SHOW format, omitting the lower body in a seated position to ensure a fair comparison. We do not include DiffSHEG~\cite{chen2024diffsheg} in our comparison since we fail to correctly visualize the SHOW results.

\subsection{Verification on DU-Trans}

\subsubsection{Quantitative Results}
We first consider a clear training set, Neutral video clips of Scott from BEAT2. 
We show that our method, trained from scratch (indicated by *), can achieve superior performance than the competing methods.
As shown in the BEAT2(Scott)-Neutral part of Tab.~\ref{tab:sota}, we begin by analyzing the synchronization of facial expressions, our method significantly outperforms the baselines in both MSE and LVD by a large margin, applicable to both specialized talking head generation methods~\cite{xing2023codetalker, fan2022faceformer} and other holistic motion generation methods~\cite{habibie2021learning, yi2023generating, liu2024emage}. 
Then, we shift our focus to the quality of gestures, where previous co-speech methods~\cite{liu2022beat, yang2023diffusestylegesture} struggle to match our results. 
Moreover, our method also outperforms TalkSHOW and EMAGE in terms of FGD and BC, with FGD being particularly notable, which indicates that the distribution of our results is the closest to the ground truth. 
Note that our method's DIV metric is slightly lower than that of TalkSHOW and EMAGE, which can be attributed to the fact that the comparative methods include some sudden and exaggerated meaningless movements, leading to a higher score. 
Finally, we evaluate the motion produced by holistic methods from an comprehensive viewpoint.
Our approach shows remarkable advancements in FMD, indicating that it excels in generating synchronized and coordinated holistic body motions. Besides, we present quantitative evaluations on Beat2-Nidal, an identity that is  distinctly different from Scott. Our method consistently verifies superior performance compared to the recent TalkSHOW and EMAGE, as shown in the BEAT2(Nidal) part of Tab.~\ref{tab:sota}.

Overall, thanks to the divide and unite mechanism in DU-Trans, our method not only demonstrates superior performance in the metrics of individual components but also clearly leads to holistic metrics. Similar observations could also be concluded from the quantitative results on SHOW dataset, as shown in Tab.~\ref{tab:sota2}. 
Notably, the solid pre-training performance brought by this powerful basic architecture will benefit the subsequent fine-tuning stage. For more details, please refer to Sec.~\ref{sec:4.3}.

\begin{table}[t]
   \caption{\textbf{Quantitative comparison with SOTA methods on SHOW dataset.} The * indicates training from scratch (pretraining).}
  \centering
  \scriptsize
   \renewcommand\arraystretch{1.2}
   \setlength\tabcolsep{6pt}
   \begin{tabular}{C{10pt}C{52pt}C{15pt}C{15pt}C{15pt}C{15pt}C{15pt}C{15pt}}
      \toprule
       Dataset &Method  & FMD$\downarrow$  & FGD$\downarrow$    & BC$\uparrow$  & DIV$\uparrow$  & MSE$\downarrow$ & LVD$\downarrow$ \\
      \midrule
      \multirow{8}{*}{{\rotatebox[origin=c]{90}{SHOW}}}
      & FaceFormer~\cite{fan2022faceformer} & -  & - & - & - & \underline{137.7} & \underline{43.86}\\
      & CodeTalker~\cite{xing2023codetalker} & - & - & - & - & 140.2 & 45.54\\
      & CaMN~\cite{liu2022beat} & 3.187 & 2.128 & 0.8035 & 9.98 & - & -\\
      & DSG~\cite{yang2023diffusestylegesture} & 3.253 & 2.349 & 0.8266 & 10.12 & - & -\\
      & Habibie \emph{et al.}~\cite{ahuja2020style} & 3.440 & 2.643 & 0.8415 & 8.007 & 145.1 & 47.11\\
      & TalkSHOW~\cite{yi2023generating} & \underline{3.038} & \underline{2.107} & \underline{0.8590} & \underline{10.25} & 139.3 & 44.81 \\
      & ProbTalk~\cite{liu2024towards} & 3.274 & 2.139 & 0.8505 & 10.04 & 151.7 & 49.66\\
      \cmidrule(lr){2-8}
      & Ours* & \textbf{2.961} & \textbf{2.002} & \textbf{0.8619} & \textbf{10.74} & \textbf{136.0} & \textbf{40.92}\\
      \bottomrule
   \end{tabular}
   \label{tab:sota2}
\end{table}

\subsubsection{Qualitative Results}
We refer readers to our project website \href{https://xc-csc101.github.io/combo/}{Combo} to for more comprehensive video demonstrations. 
In this section, we focus qualitative comparison with two recent SOTA methods, TalkSHOW and EMAGE, which can generate coherent motions. While the previous approaches like Habibie \etal suffer from varying degrees of jittering, thus we do not include them in this comparison. 

On the BEAT2 dataset, we focus on the Scott-Neutral part. As depicted in Fig.~\ref{fig:sota_beat}, TalkSHOW typically displays much slower and less varied motion than the other two, regardless of how the rhythm and semantic content of the audio change. 
EMAGE, as an improved version of TalkSHOW, utilizes a more comprehensive VQ-VAE to encode the full body and also meticulously upgrades the motion generation network, resulting in superior performance compared to the former. 
For example, EMAGE can respond to certain words that serve as emphasis, such as "best" and "leading". 
Yet, it still fails to generate expressive and lifelike gestures. 
In contrast, our method is not only effective in conveying some concepts such as "Buddhist temple" but also in interpreting semantic words like "includes" and "come to", while being synchronized with rhythmic elements and emphasis, such as "best" and "leading". 
We further attach the facial meshes to evaluate the lip synchronization at the right part. 
In comparison with other methods, our model produces expressions that are accompanied by more precise lip shapes and more natural facial motions. 

On the SHOW dataset~\cite{yi2023generating}, the qualitative results are shown in Fig.~\ref{fig:sota_show}. 
Undoubtedly, our method excels in other methods mainly in three aspects: semantic and gesture alignment, rhythm and gesture consistency, and content and facial expression synchronization. 
Additionally, the coordination among various body parts is not easily depicted in Figs.~\ref{fig:sota_beat} and~\ref{fig:sota_show}, we further provide evaluation results in subsequent user studies. 

\subsubsection{User Study}
\label{sec:4.2.3}
It is widely acknowledged that assessing the quality of generative tasks is inherently subjective. Despite that we have evaluated with multiple quantitative metrics, there remains a significant
gap between such and human visual perception~\cite{GENEA, QPGESTURE}. To this end, we conduct rigorous user studies to compare our framework with three baselines: mocap recordings (GT), TalkSHOW~\cite{yi2023generating}, and EMAGE~\cite{liu2024emage} on BEAT2 dataset.

Specifically, we utilize the official code from EMAGE to visualize the motions. To eliminate potential human bias, all rendered videos are generated using the fixed body shape. All segments are approximately 6 to 8 seconds long and should not be shorter than 5 seconds. They are complete phrases, beginning and ending at word boundaries, and do not end abruptly or on a “cliffhanger”. We recruit 40 participants from our lab\footnote{https://april.zju.edu.cn/}, all of whom have strong English proficiency and are ensured to be with no prior knowledge/awareness of this project before the user study. The majority are male (about 75\%), with the remaining being female. Almost all participants are residents of mainland China, except for one international student from South Korea. Their ages range from 20 to 30 years old. Each participant receive a 9 GBP compensation, determined based on the average completion time (about 45 minutes) of our user study (including Sec.~\ref{sec:4.2.3}, Sec.~\ref{sec:4.3.4}, and Sec.~\ref{sec:4.4.3}) and the average wage level~\cite{GENEA}.

All tests are conducted using \textit{pairwise} comparisons. Participants are shown two video clips synthesized by different models on each page. They are then required to label their preferred clip based on the instructions displayed below the videos. Besides, each participant is required to complete a substantial \textit{training page} featuring a fixed set of videos before starting each test, to familiarize them with the task. Moreover, an \textit{attention check} is randomly inserted into the experiment to ensure valid responses. Concretely, a text message reading: ``Attention: Please select the right motion" appear both at the bottom of the video pair for the entire duration of the question and during the transition gap between the two clips. Any samples that fail the attention check are excluded from the final analysis. Another filtering is also employed, users are asked to provide a brief qualitative reason for their choice in a random page of each test. This written feedback helps us assess the reliability of the results.

After the above preliminaries, we follow the evaluation in GENEA~\cite{GENEA} and propose three types of preference tests: one test assesses \textit{human-likeness}, while the other two evaluate \textit{appropriateness} in different aspects: speech-motion synchronization, face-body coordination. We report \textit{pairwise wining proportion} in the human-likeness test and \textit{matched percentage} in the appropriateness test. We adopt the pre-trained Combo on neutral data in these evaluations.

\begin{figure}[t!]
	\centering
	\includegraphics[width=0.48\textwidth]{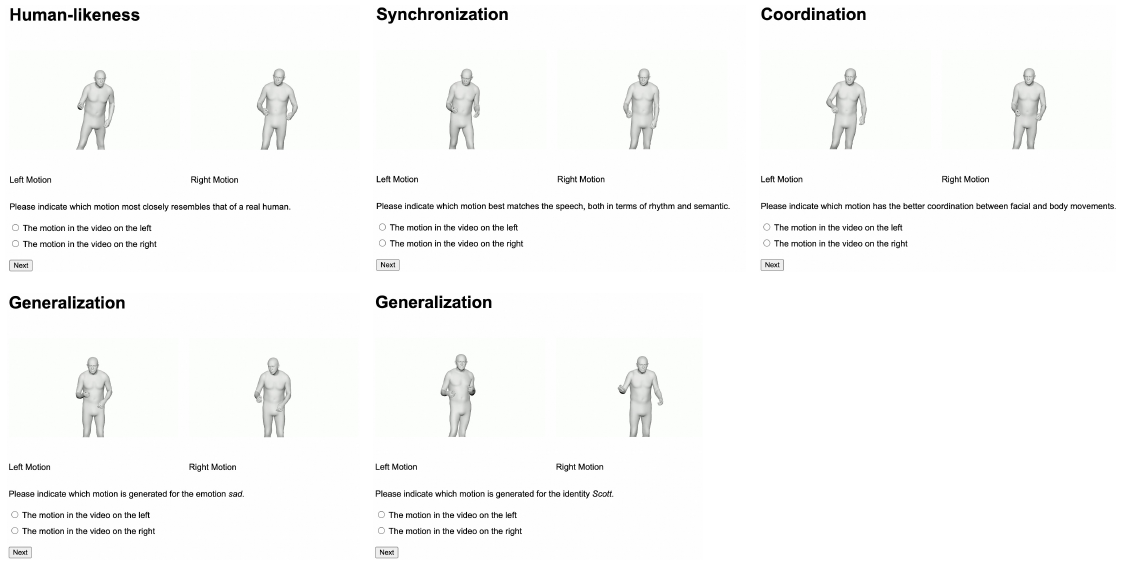}
	\caption{\textbf{Five screenshots of the user interface from our studies.} Please zoom in for more details.}
	\label{fig:screeshot}
\end{figure}

\begin{table}[t!]
\caption{\textbf{Average scores of user study with 95\% confidence intervals.} In each test, GT serves as the topline, with higher scores indicating better performance across all metrics. We omit the \%. Among all tests, TalkSHOW and EMAGE involve a single model, whereas our method involves both pretraining and finetuning versions.}
  \centering
  \scriptsize
   \renewcommand\arraystretch{1.2}
   \setlength\tabcolsep{1.9pt}
   \begin{tabular}
    {C{35pt}C{46pt}C{37pt}C{36pt}C{42pt}C{39pt}}
      \toprule
         & Human-likeness  & Synchronized & Coordinated & Generalized-E  & Generalized-I \\
      \midrule
      GT & 86.9$\pm$4.2 & 79.5$\pm$4.0   & 80.3$\pm$3.8 & 78.9$\pm$3.6 & 98.4$\pm$5.3 \\
      TalkSHOW~\cite{yi2023generating} & 15.2$\pm$4.6  & 53.5$\pm$3.1 & 52.0$\pm$2.4 & 58.7$\pm$5.1 & 51.7$\pm$2.2\\
      EMAGE~\cite{liu2024emage} & 43.4$\pm$5.6  & 57.7$\pm$3.0 & 59.8$\pm$4.3 & 61.2$\pm$4.9 & 53.9$\pm$3.2\\
      Ours & 58.2$\pm$6.1   & 70.8$\pm$4.2 & 68.4$\pm$4.7 & 73.1$\pm$3.6 & 96.1$\pm$5.0\\
      \bottomrule
   \end{tabular}
   \label{tab:user}
\end{table}

\noindent\textbf{Human-likeness.}
In the human-likeness test, participants are asked to evaluate ``Please indicate which motion most closely resembles that of a real human". For this test, we select 24 speech segments from the BEAT dataset's test set of Speaker-2 (BEAT-S2 is chosen due to baseline models availability for this identity) to generate corresponding motions, resulting in 24 video clips for each method. These video clips are muted to eliminate any influence from the speech content. Four approaches yields 12 possible pairwise combinations for side-by-side comparisons. Each participant assess 24 video pairs, covering all 24 speech samples, with each of the 12 combinations appearing twice. The speech samples and the pairing of comparisons are randomized for each participant.

In this experiment, 40 participants pass the filtering checks, and the results are presented in Tab.~\ref{tab:user} column 1. Our method outperforms the two recent baselines, which verifies that our method significantly surpasses the compared SOTA methods of human-likeness.

\noindent\textbf{Synchronization.}
In the synchronization test, participants are asked to ``Please indicate which motion best matches the speech, both in terms of rhythm and semantic”. We follow \textit{mismatching} methodology as GENEA~\cite{GENEA}. Specifically, we select 10 speech segments from BEAT-S2 to create 10 video clips per method. We then rearrange the speech-motion segments so that each motion is paired with an unsynchronized speech. This creates a pair of motion videos features the \textit{same} speech audio, one matched and one mismatched. Notably, mismatched pairs may have misaligned durations. We address this using the method used in GENEA. The position of the mismatched video (left or right) is randomized. Each participant is asked to assess 40 pairs (10 speech segments $\times$ 4 approaches).

In this experiment, 37 subjects pass the filtering checks, and the results are shown in Tab.~\ref{tab:user} column 2. 
Matched percentage identities how often subjects prefer matched over mismatched motion. Because the higher the motion-speech synchronization in the generated results, the more easily participants can detect mismatches. Consequently, our method is more synchronized for the specific speech signal compared to other recent methods.

\noindent\textbf{Coordination.}
The coordination test is largely consistent with the synchronization test, with the only difference lying in the construction of the mismatched pairs. Participants are asked to "Please indicate which motion has the better coordination between facial and body movements". They are shown a pair of videos: one shows facial and body movements generated from the same speech segment, and the other are from different samples. The videos are muted in this test. In this experiment, 39 subjects pass the filtering checks and the results are shown in Tab.~\ref{tab:user} column 3. The mismatched percentage shows how often subjects notice that the face and body motions come from different samples. The more harmonious the face and body motions are, the higher the rate at which mismatches are identified after being shuffled. The results verify that the body and face movements generated by our method are better aligned with each other due to the divide-and-unite design.

\subsection{Verification on \texttt{X}-Adapter}
\label{sec:4.3}
In Sec.~\ref{sec:4.3.1}, we first analyze the effectiveness of the \texttt{X}-adapter in emotion transfer during training, and then further explore the two benefits it brings: multi-modal editing and unseen emotion editing during inference. These experiments are conducted on Scott data. Then, in Sec.~\ref{sec:4.3.2}, we analyze the effectiveness of the \texttt{X}-adapter in identity transfer and validate cross-identity transfer under various conditions. These experiments are conducted on Neutral data. 
We further conduct the user study to supplement more comprehensive subjective analysis in Sec.~\ref{sec:4.3.4}. Finally, we provide a detailed analysis of overall tuning efficiency in Sec.~\ref{sec:4.3.3}.
\begin{figure}[t!]
	\centering
	\includegraphics[width=0.48\textwidth]{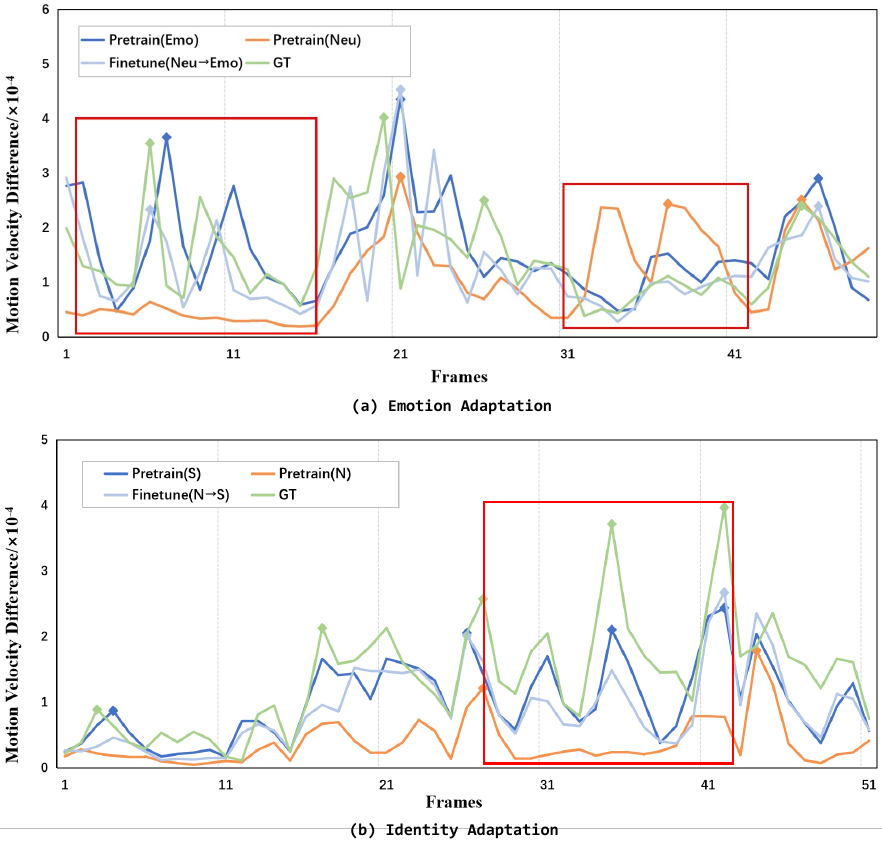}
	\caption{\textbf{Velocity comparisons on BEAT2 of identity and emotion transfer.} Velocity is calculated as the frame-by-frame channel average of the absolute residuals of motion in adjacent frames. These visualizations verify the effectiveness of \texttt{X}-Adapter on sub-task finetuning. Please pay attention to the area highlighted in red rectangular.}
	\label{fig:analysis}
\end{figure}

\begin{table}[t]
   \caption{\textbf{Quantitative comparison with several variants of emotion transfer on Scott data.} Emo means emotional while Neu means neutral. The "$\rightarrow$" means transferring the pretrained model on source data to the target by fast finetuning. Each row represents the test results of the corresponding method on Scott-Emotional data.
}
  \centering
  \scriptsize
   \renewcommand\arraystretch{1.2}
   \setlength\tabcolsep{6pt}
   \begin{tabular}{C{2pt}C{58pt}C{15pt}C{15pt}C{15pt}C{15pt}C{15pt}C{15pt}}
      \toprule
       Data &Method  & FMD$\downarrow$  & FGD$\downarrow$    & BC$\uparrow$  & DIV$\uparrow$  & MSE$\downarrow$ & LVD$\downarrow$ \\
      \midrule
      \multirow{3}{*}{{\rotatebox[origin=c]{90}{Emotional}}}
      &Pretrain(Neu) & 1.749 & 1.403 & 0.7810 & 9.257 & 5.551 & 6.880\\
      &Pretrain(Emo) & \textbf{1.120} & 0.599 & \textbf{0.8064} & \textbf{11.08} & 5.149 & 6.434\\
      &Finetune(Neu$\rightarrow$Emo) & 1.128 & \textbf{0.568} & 0.8003 & 10.63 & \textbf{5.015} & \textbf{6.408}\\
      \bottomrule
   \end{tabular}
   \label{tab:emo}
\end{table}

\begin{figure*}[t!]
	\centering
	\includegraphics[width=0.95\textwidth]{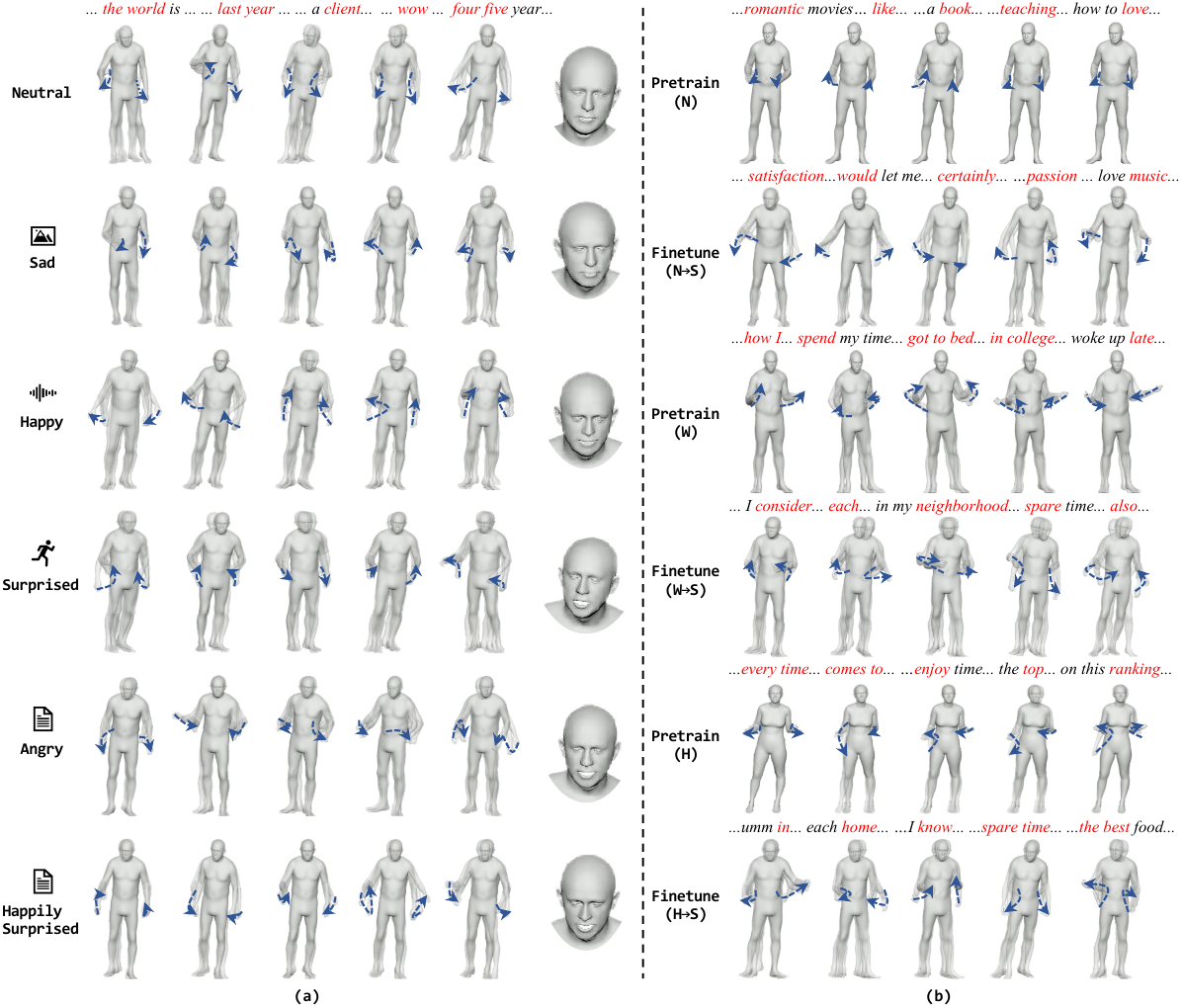}
	\caption{\textbf{The visualization of multi-modal emotion control and different identity personalization.} The left part shows the manipulated outputs guided by sad images, happy audio, surprised motion clips, and angry text. The bottom row is the result of an unseen emotion represented by a happily surprised text prompt. The right part displays the motions from the various source identities as well as the motions after fine-tuning that transfers them to the target identity Scott.}
	\label{fig:analysis2}
\end{figure*}

\subsubsection{Analysis on Emotion Transfer}
\label{sec:4.3.1}
\noindent\textbf{The Effectiveness of Emotion Transfer.}
Our proposed \texttt{X}-Adapter enables the transfer from the emotion-agnostic talking body models into the emotional ones, we first give a visualization of motion velocity in Fig.~\ref{fig:analysis}(a) to verify the effectiveness of its emotion transfer ability. 
Specifically, we train two models with our framework: 1) trained with neutral data via DU-Trans; 2) pretrained with neutral data via DU-Trans and then finetuned with emotional data with our \texttt{X}-Adapter. 
During the test, we randomly sample an emotional clip from the Scott test splits and visualize the velocities of the motions generated by three models, and compare them to the ground truth. 
We observe that the neutral model outputs a limited dynamic and varied motion, but after fine-tuning, it aligns closely with the motion patterns of the ground truth, indicating that the emotion transfer is quite effective. 
Then, we report the quantitative results in Tab.~\ref{tab:emo} consistent with the aforementioned conclusions, models from the neutral domain do not perform well on emotional data, as evidenced by their lower performance on all metrics at row Pretrain(Neu), and conversely, after fine-tuning, the performance has seen an overall enhancement as shown in row Finetune(Neu$\rightarrow$Emo). To verify the SOTA level of our fine-tuning results, we copy the values of Finetune(Neu$\rightarrow$Emo) in Tab.~\ref{tab:sota} BEAT2(Scott)-Emotional part marked as Ours$\dagger$. Our method significantly outperforms the baselines on almost all metrics, which is attributed to the solid pretraining performance and excellent cross-domain adaptability. 

Additionally, we present the outputs of DU-Trans when fully trained from scratch on emotional data. The results are shown in Fig.~\ref{fig:analysis} at curve Pretrain(emo) and Tab.~\ref{tab:emo} at row Pretrain(emo), both also exhibit remarkable performance. The comparable
results of row Pretrain(emo) and row Finetune(Neu$\rightarrow$Emo) indirectly demonstrates the
superiority of DU-Trans’s design. However, this manner lacks flexible editing and adaptability, like the current baselines~\cite{yi2023generating, liu2024emage, liu2024towards, chen2024diffsheg}.

\noindent\textbf{Multi-Modal Emotion Editing.}
As mentioned in Sec.~\ref{sec:edit}, our framework allows multi-modal emotion editing. 
As shown in Fig.~\ref{fig:analysis2}(a), given neutral audio and multi-modal emotional conditions, we consistently generate gestures that accurately reflect the emotional cues contained in various guidance, including image, audio, motion sequence, and text. 
For example, our method produces lowered hand movements that are associated with the sadness depicted in a sad image, \eg, graves, whereas it generates a variety of large gestures with rhythmic body swaying when given a happy audio. 
Similarly, our method also successfully captures key actions from the surprise motion sequence, such as more frequent body turns and hurried upward gestures. 
For angry text, it includes more abrupt downward pressing gestures. 
Additionally, in the fifth column, we display facial meshes at a certain word under different emotions. 
It is evident that emotions do not interfere with the articulation of speech content, as can be seen from the relatively consistent mouth shapes, but they do influence the overall expression. 
For instance, when happy, the corners of the mouth turn upwards, and when surprised, the mouth opens wide. 
Thus, our method allows for precise emotion control of motion generation through any modality, flexibly supporting editing needs in various situations.

\noindent\textbf{Unseen Emotion Editing.}
In addition to the above, we supplement a qualitative study to demonstrate that the emotion space under the CLIP domain possesses a certain degree of generalization ability, thanks to the rich semantics inherited from CLIP. As shown in Fig.~\ref{fig:analysis2}(a), row 6 shows the results of the given \textit{happily surprised} in text prompts. 
This is an unseen compound emotion, and the generated motions accurately convey happiness and surprise simultaneously, which can be observed from the overall lively body movements and astonished expressions. Besides, by comparing the facial details with those of happy (row 3) or surprised (row 4), the result of this unseen style is not a mere replication of either but rather a full combination of both. This experiment verifies the flexibility and rich semantic priors of the CLIP feature space.

\subsubsection{Analysis on Identity Transfer}
\label{sec:4.3.2}
\noindent\textbf{The Effectiveness of Identity Transfer.}
Similar to Sec.~\ref{sec:4.3.1}, we perform qualitative and quantitative experiments to verify the effectiveness of \texttt{X}-Adapter in identity transfer. 
First, we visualize the motion velocity curves for four methods, \ie, ground truth, Scott model, Nidal model, and the Nidal model after finetuned on Scott data, when each receives a randomly sampled Scott audio signal. 
As shown in Fig.~\ref{fig:analysis}(b), Nidal usually speaks in a calm and quiet manner, which corresponds to the motion curve displaying a smooth low-amplitude trajectory, as colored in orange. 
On the other hand, our fast adaption helps to transfer to the lively and dynamic Scott style, \ie, the light blue curve exhibits similar variations and intensities to the ground truth in green and the Scott model in blue. 
Second, we attach the quantitative results in Tab.~\ref{tab:id} to support the above observations. Concretely, row Pretrain(N) demonstrates that direct cross-identity evaluation does not yield satisfactory results, while slight fine-tuning can achieve competitive performance compared to a model sufficiently trained from scratch, as shown by comparing row Pretrain(S) and row Finetune(N$\rightarrow$S). 

\begin{table}[t!]
   \caption{\textbf{Quantitative comparison with several variants of identity transfer on neutral data.} S, N, W, and H are the abbreviations for Scott, Nidal, Wayne, and Hailing, respectively. Each row represents the test results of the corresponding method on Scott-Neutral data.}
  \centering
  \scriptsize
   \renewcommand\arraystretch{1.2}
   \setlength\tabcolsep{6pt}
   \begin{tabular}{C{2pt}C{52pt}C{15pt}C{15pt}C{15pt}C{15pt}C{15pt}C{15pt}}
      \toprule
       Data &Method  & FMD$\downarrow$  & FGD$\downarrow$    & BC$\uparrow$  & DIV$\uparrow$  & MSE$\downarrow$ & LVD$\downarrow$ \\
      \midrule
      \multirow{7}{*}{{\rotatebox[origin=c]{90}{Scott-Neutral}}}
      &Pretrain(S) & \textbf{1.098} & 0.563 & 0.8023 & \textbf{10.48} & 5.098 & 6.005\\
      \cmidrule{2-8}
      &Pretrain(N) & 9.231 & 2.883 & 0.0149 & 0.551 & 15.20 & 11.43\\
      &Finetune(N$\rightarrow$S) & 1.326 & 0.680 & 0.7888 & 9.790 & \textbf{4.926} & \textbf{6.002}\\
      &Pretrain(W) & 3.955 & 3.656 & 0.4379 & 3.265 & 8.017 & 7.507\\
      &Finetune(W$\rightarrow$S) & 1.223 & 0.513 & 0.8083 & 9.998 & 5.800 & 6.431\\
      &Pretrain(H) & 7.752 & 6.488 & 0.6393 & 5.691 & 10.76 & 9.243\\
      &Finetune(H$\rightarrow$S) & 1.137 & \textbf{0.480} & \textbf{0.8169} & 10.02 & 5.860 & 6.614\\
      \bottomrule
   \end{tabular}
   \label{tab:id}
\end{table}

\noindent\textbf{Cross-Identity Transfer Analysis.}
To further verify that our method is robust to the choice of source identity and can be transferred to any other identity, we provide a cross-identity transfer analysis. Specifically, we construct three identity pairs with significant gaps: 1) Nidal and Scott have different personalities, one being quiet and the other being lively; 2) Wayne and Scott, although similar in temperament, have different behavioral habits; 3) Hailing and Scott have different genders. 
As shown in Tab.~\ref{tab:id}, the rows Pretrain(S), Finetune(N$\rightarrow$S), Finetune(W$\rightarrow$S), and Finetune(H$\rightarrow$S) exhibit a comparable performance, from which we can infer that the above three distinct identities can all be transferred onto Scott style. 
Besides, we supplement the intuitive visualizations in Fig.~\ref{fig:analysis2}(b). 
In line with the above, each person exhibits their own unique movement dynamics. 
Nidal (row 1) shows less variety in his movements and changes slowly. Wayne (row 3) is accustomed to swinging his arms and speaking while facing to the left. Hailing (row 5) possesses a characteristically feminine grace. 
After efficient finetuning (rows 2, 4, 6), our approach successfully adjusts the movement patterns to match Scott's, no matter the initial state. 
In conclusion, \texttt{X}-Aapter is capable of handling various challenging identity transfer tasks.

\subsubsection{User Study}
\label{sec:4.3.4}
Similarly, we use the same methodology in Sec.~\ref{sec:4.2.3} and conduct the appropriateness tests following the mismatching strategy to verify the effectiveness of \texttt{X}-Adapter on emotion and identity transfer. Our method for this evaluation is the finetuning version. We also include TalkSHOW and EMAGE in the test to demonstrate not only that Combo can generalize to new speakers and emotions, but also that these other models cannot.

For emotion, we sample 2 clips from each emotional speech segment of BEAT-S2, obtaining 14 motion videos (2 segments $\times$ 7 emotion styles). Each is paired with a video generated from a neutral speech segment to form a mismatched pairs. In the emotion generalization test, participants are asked to ``Please indicate which motion is generated for the given emotion" (covering 7 emotional styles, excluding neutral).

For identity, we use 4 identities (Speaker-1, -2, -11, -23). Each identity forms 3 mismatched pairs with the remaining identities, such as 1-2, 1-11, 1-23, where the first number represents the given identity in question. This allows us to construct 12 combinations (4 identities $\times$ 3 pairs), with each identity is included 6 times. 
Concretely, we sample 6 speech segments from each identity, with each of the 12 combinations appearing once. In this test, participants are asked to ``Please indicate which motion is generated for the given identity".

Notably, the videos in both emotion and identity experiments are muted. Besides, the training pages in the experiments include not only the fixed pair videos but also example videos for the 7 emotions and 4 identities that users need to identify.
         
As a result, 38 and 39 subjects pass the filtering checks, respectively. The results are shown in Tab.~\ref{tab:user} columns 4 and 5.
For baselines like EMAGE, the officially released model has relatively weak emotion control, which is realized through the input of emotional speech segments. Its ability to transfer speaker-ID is virtually non-existent due to identity-specific training. The above description can be consistently inferred from their low scores, indicating that their motions across different emotions and identities are difficult to distinguish. Conversely, our results are easily identifiable, demonstrating that our method can generalize to new speakers and emotions, which is attributed to the fast-finetuning property of the proposed \texttt{X}-Adapter.

\begin{figure}[t!]
	\centering
	\includegraphics[width=0.48\textwidth]{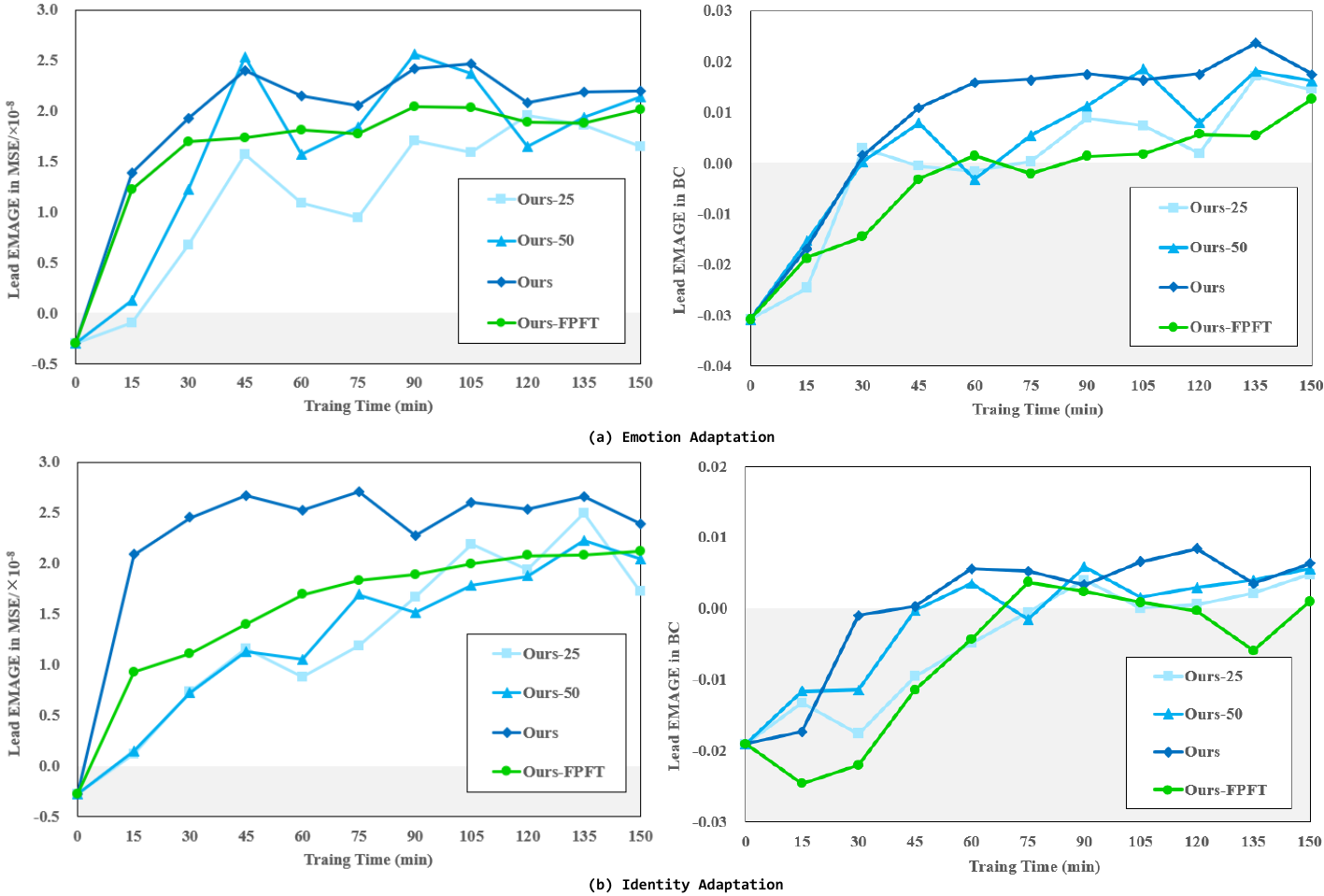}
	\caption{\textbf{Tuning efficiency of \texttt{X}-Adapter}. In this visualization, we choose MSE for the face and BC for the body. Values below 0 on the y-axis (gray fill) indicate inferior performance compared to EMAGE, and vice versa. Our design exhibits exceptional tuning efficiency in terms of training time and data, achieving SOTA performance within 45 minutes with full (Ours) or half data (Ours-50), or even within 90 minutes with only 25\% training data (Ours-25). Ours-FPFT means full-parameter finetuning on DU-Trans (w/o. \texttt{X}-adapter) with full data.}
	\label{fig:efficiency}
\end{figure}

\subsubsection{Tuning Efficiency}
\label{sec:4.3.3}
In this part, we provide a detailed analysis to demonstrate that \texttt{X}-Adapter can efficiently adapt the pretrained DU-Trans of Scott neutral data to alternative identity or emotion, even with limited training data. 

Firstly, we emphasize that our finetuning only updates about 4M parameters for both identity and emotion adaptation, in which the adapter is 3.73M and three heads are 0.27 M, while the total number of parameters is around 40M, meaning we only utilize 10\% of the trainable parameters yet achieve superior performance.

Secondly, we analyze the efficiency of our finetuning on training \emph{time} and \emph{data}. 
Specifically, we conduct periodic tests every 15 minutes during the identity and emotion transfer process to evaluate the tuning time and data efficiency and record the MSE metric for face expression and BC metric for body gesture at each time point. 
Comparing our method to EMAGE, we plot the leading value in Fig.~\ref{fig:efficiency}. 
Specifically, our method outperforms SOTA results within 45 minutes with full or half data, and can also achieve the best output with just a quarter of the data within 90 minutes. 
Notably, we open all parameters of DU-Trans and use full data for finetuning (as shown in green color), but its performance is not as good as parameter-efficient finetuning (as shown in deep blue color), even costing many trainable parameters.
Beyond the above, we find that: 1) the time required for facial performance to reach comparable levels is relatively shorter than that for gestures, as expressions only involve a small range of facial movements, whereas gestures encompass a larger scope. For example, just 15 minutes of fine-tuning is sufficient for our method to surpass the SOTA competitor on the MSE metric; 
2) in contrast to finetuning with a limited dataset, employing the full dataset can swiftly improve performance in a short period, but as time progresses, it typically saturates at a comparable level in the end. 
To sum up, users can dynamically adjust the finetuning duration based on the amount of data available.

\begin{table*}
\caption{\textbf{Quantitative ablations for DU-Trans and \texttt{X}-Adapter.} The ablation studies of DU-Trans are conducted on the Scott-Neutral in BEAT2 while that of \texttt{X}-Adapter are conducted on the Scott-Emotional.}
    \centering
    \scriptsize
    \resizebox{1\linewidth}{!}
    {
    \begin{subtable}[htbp]{0.48\linewidth} 
        \centering
        \scriptsize
        \renewcommand\arraystretch{1.1}
         \setlength{\tabcolsep}{6pt} 
        \caption{Ablation on DU-Trans.}
        \begin{tabular}{C{1pt}C{40pt}C{10pt}C{15pt}C{15pt}C{15pt}C{15pt}C{15pt}C{15pt}}
      \toprule
        &Method  & P.(M) & FMD$\downarrow$ & FGD$\downarrow$    & BC$\uparrow$  & DIV$\uparrow$  & MSE$\downarrow$ & LVD$\downarrow$ \\
      \midrule
      \multirow{5}{*}{{\rotatebox[origin=c]{90}{Architecture}}}
       & $l$ = 0, $j$ = 8 & 19.46 & 1.265 & 0.635 & 0.7921 & 9.972 & 5.795 & 6.611\\
       & $l$ = 8, $j$ = 0 & 35.89 & 1.336 & 0.588 & 0.7695 & 9.350 & 5.465 & 6.428\\
       & $l$ = 7, $j$ = 1 & 34.84 & \textbf{1.190} & \textbf{0.579} & \textbf{0.7999} & \textbf{9.991} & \textbf{5.362} & \textbf{6.358}\\
       & $l$ = 5, $j$ = 3 & 32.74 & 1.214  & 0.580  & 0.7904 & 9.895  & 5.488  & 6.529\\
       & $l$ = 3, $j$ = 5 & 30.64 & 1.278  & 0.589  & 0.7917 & 9.658  & 5.602  & 6.601\\
       \cmidrule(lr){2-9}
       \multirow{4}{*}{{\rotatebox[origin=c]{90}{Loss}}}
       & w/o. $\mathcal{L}_F$,$\mathcal{L}_B$ & 34.84 & 30.57 & 27.19 & 0.8442 & 32.46 & 2910 & 109.3\\
       & $\lambda_F$,$\lambda_B$ = 1 & 34.84 & 1.412  & 0.593  & 0.7978 & 9.963  & 5.373  & 6.407\\
       & $\lambda_F$,$\lambda_B$ = 0.5 & 34.84 & \textbf{1.190} & \textbf{0.579} & \textbf{0.7999} & \textbf{9.995} & \textbf{5.362} & \textbf{6.358}\\
       & $\lambda_F$,$\lambda_B$ = 0.1 & 34.84 & 1.468  & 0.598  & 0.7904 & 9.880  & 5.671  & 6.544\\
       \cmidrule(lr){2-9}
       \multirow{5}{*}{{\rotatebox[origin=c]{90}{Bi-Flow}}}
       & $l$ = 1 & 36.94 & 1.198 & 0.586 & 0.7927 & 9.717 & 5.226 & 6.299\\
       & $l$ = 3 & 36.94  & 1.098 & \textbf{0.563}  & 0.8023  & \textbf{10.48}  & 5.098  & 6.005\\
       & $l$ = 3,Uni-BF & 35.90  & 1.126 & 0.572  & 0.8011 & 10.14 & 5.159  & 6.087  \\
       & $l$ = 3,Uni-FB & 35.90  & 1.119 & 0.567  & 0.8020  & 10.37 & 5.285  & 6.286\\
       & $l$ = 6 & 36.94  & 1.128 & 0.579  & 0.8002 & 9.986  & 5.212  & 6.010\\
       & $l$ = 3,4 & 39.05 &\textbf{1.097} & 0.565  & \textbf{0.8025}  & 10.45 & 5.111 & 6.001\\
       & $l$ = 2,3,4 & 41.15 & 1.105  & 0.568  & 0.8020 & 10.44  & \textbf{5.087}  & \textbf{5.997}\\
      \bottomrule
   \end{tabular}
    \end{subtable}

    \begin{subtable}[htbp]{0.48\linewidth} 
        \centering
        \scriptsize
        \renewcommand\arraystretch{1.2}
         \setlength{\tabcolsep}{6pt} 
        \caption{Ablation on X-Adapter.}
        \begin{tabular}{C{1pt}C{39pt}C{10pt}C{15pt}C{15pt}C{15pt}C{15pt}C{15pt}C{15pt}}
      \toprule
        &Method  & P.(M) & FMD$\downarrow$  & FGD$\downarrow$    & BC$\uparrow$  & DIV$\uparrow$  & MSE$\downarrow$ & LVD$\downarrow$ \\
      \midrule
      \multirow{4}{*}{{\rotatebox[origin=c]{90}{Location}}}
       & Serial & 4.00 & 1.332 & 0.595 & 0.7793 & 10.58 & 5.610 & 6.840\\
       & Parallel & 4.00 & \textbf{1.128} & \textbf{0.568} & \textbf{0.8003} & \textbf{10.63} & \textbf{5.015} & \textbf{6.408}\\
       & Only MHA & 2.14 & 1.312 & 0.581 & 0.7942 & 10.63 & 5.237 & 6.581\\
       & Only FFN & 2.14 & 1.427 & 0.595 & 0.7825 & 10.60 & 5.282 & 6.554\\
       \cmidrule(lr){2-9}
       \multirow{3}{*}{{\rotatebox[origin=c]{90}{Condition}}}
       & w/o. $\texttt{x}$ & 4.00 & 1.236 & 0.572 & 0.7953 & 10.11 & 5.329 & 6.672\\
       & Stylization & 8.14 & \textbf{1.120} & \textbf{0.567} & 0.8002 & 10.59 & 5.228 & 6.506\\
       & Add & 4.00 & 1.128 & 0.568 & \textbf{0.8003} & \textbf{10.63} & \textbf{5.015} & \textbf{6.408} \\
       \cmidrule(lr){2-9}
       \multirow{3}{*}{{\rotatebox[origin=c]{90}{Scale}}}
        & Scalar-1.0 & 3.99 & 1.236 & 0.569 & 0.7987 & 9.890 & 7.044 & 7.316\\
       & L-Scalar & 3.99 & 1.256 & 0.584 & 0.8001 & 10.15 & 5.464 & 6.781\\
       & Dy-Scale & 4.00 & \textbf{1.128} & \textbf{0.568} & \textbf{0.8003} & \textbf{10.63} & \textbf{5.015} & \textbf{6.408}\\
       \cmidrule(lr){2-9}
       \multirow{4}{*}{{\rotatebox[origin=c]{90}{PEFT}}}
        & LoRA-r64 & 2.17  & 1.205  & 0.581  & 0.7911  & 10.17  & 5.341  &  6.659\\
       & Prefix Tuning & 0.03 & 2.234 & 1.281 & 0.7673 & 6.156 & 7.943 & 8.063\\
       & Adapter-r64 & 2.17 & 1.191 & 0.578 & 0.7906 & 10.37 & 5.226 & 6.654\\
       & Adapter-r128 & 4.00 & \textbf{1.128} & \textbf{0.568} & \textbf{0.8003} & \textbf{10.63} & \textbf{5.015} & \textbf{6.408}\\
      \bottomrule
   \end{tabular}
    \end{subtable}
}
\label{tab:abla}
\end{table*}

\begin{figure}[t!]
	\centering
	\includegraphics[width=0.48\textwidth]{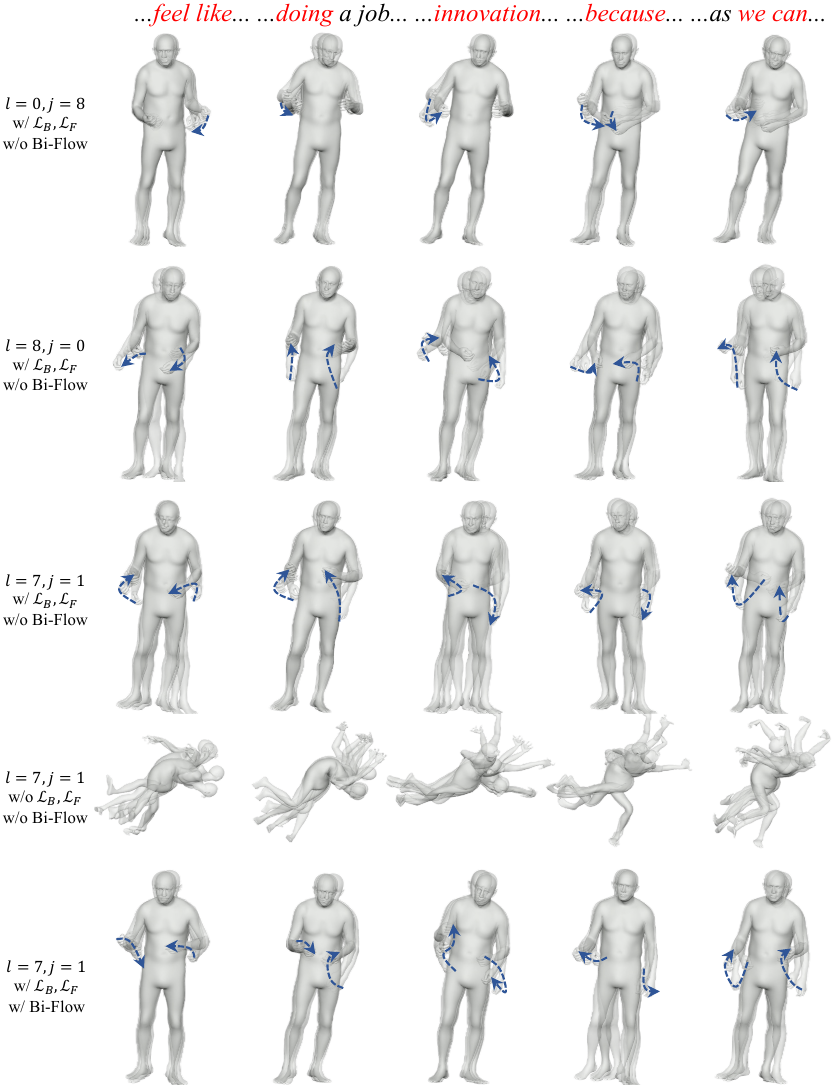}
	\caption{\textbf{Qualitative ablations for DU-Trans.} The full version of DU-Trans ($l=7,j=1,w/ \mathcal{L}_B, \mathcal{L}_F, w/ \text{Bi-Flow}$) produces shows more synchronized and meaningful motions than other ablations.}
	\label{fig:abla}
\end{figure}

\subsection{Ablation Study}
In this part, we ablate on the proposed DU-Trans and \texttt{X}-Adapter to validate our design choices and hyperparameters. 

\subsubsection{Ablation on DU-Trans}
In Tab.~\ref{tab:abla}(a), we conduct the ablation study on DU-Trans from three aspects, \ie, architecture, loss functions, and Bi-Flow variants. 
The architecture ablation study employs auxiliary losses with optimal weights but does not utilize the Bi-Flow. 
The loss ablation study fixes the optimal architecture without including the Bi-Flow. Finally, the last study ablates the Bi-Flow on the optimal architecture incorporated with auxiliary losses. 

\noindent\textbf{Analysis on Basic Architecture (Tab.~\ref{tab:abla}(a) Architecture part).}
The insight of our DU-Trans is to use divided encoders to ensure sufficient exhibition for both face and body, followed by a united decoder to implicitly model their interconnections and to directly predict the combined coefficients for overall high coordination. 
To verify its effectiveness, we first design two variants, the one with only two divided encoders (row $l=8$, $j=0$, where $l$ is the number of encoder layers and $j$ is the number of decoder layers) and the other with only a united decoder (row $l=0$, $j=8$). The results indicate that neither of them yields satisfactory outputs. 
The underlying reason may be that the two separate encoders do not account for each other's influence, which is evident from the significant degradation in FMD. 
Directly employing a single decoder may cause two distinct distributions to converge toward each other, ensuring harmony but greatly sacrificing the uniqueness of each, especially as indicated by the decline in FGD and MSE. 
We also give a visualization in Fig.~\ref{fig:abla} that the motion generated by this method has a minimal amplitude. 
Then, we explore how to preserve individual differences while achieving a harmonious commonality. 
As shown in row $l=7$, $j=1$, when the encoder has 7 layers and the decoder has 1 layer, our DU-Trans achieves the best performance. 

\noindent\textbf{Analysis on Loss Functions (Tab.~\ref{tab:abla}(a) Loss part).}
The two auxiliary losses for face and body branches are critical to distilling their respective features and aiding in the joint learning of holistic motion. 
To verify their effectiveness, we present the quantitative results in the Loss part. 
It is obvious that the absence of auxiliary loss (row w/o. $\lambda_F,\lambda_B$) leads to a significant decrease in most metrics except for BC and DIV. 
These two are unreliable when there is a noticeable jitter in the motion. 
This observation can also be discerned in the fourth row in Fig.~\ref{fig:abla}, \ie, exhibiting unreasonable poses and meaningless movements. Furthermore, we explore the optimal hyperparameters for these losses and find that the optimal set is $\lambda_F = \lambda_B = 0.5$. 

\noindent\textbf{Analysis on Bi-Flow (Tab.~\ref{tab:abla}(a) Bi-Flow part).}
The Bi-Flow design is utilized during the divide phase for the face and body to provide global dynamic cues that enhance their respective performances. 
As shown in the Bi-Flow part, we examine the impact of hyperparameters regarding Bi-Flow, including its position (first 3 rows in this part) and number (the remaining). 
Concretely, when we apply this module at the shallow layer $l=1$ and the deep layer $l=6$, the performance does not match that at the intermediate layer $l=3$. 
From row $\lambda_F = \lambda_B = 0.5$ of the Loss part and $l=3$ (including two unidirectional variants body-to-face Uni-BF and face-to-body Uni-FB) of the Bi-Flow part, we observe that the unidirectional
variants shows enhancement on the target component and overall improvement, while incorporating bidirectional interaction achieves more comprehensive gains. This demonstrates that a
bidirectional connection is superior, creating a synergistic effect where the body provides global context to the face (shown in EMAGE) and the face provides fine-grained cues back to the body (shown in DiffSHEG), leading to mutual reinforcement for holistic harmony.
We also provide a qualitative comparison in Fig.~\ref{fig:abla} to support the above conclusion. 
Furthermore, building on the foundation of $l=3$, we increase the number of Bi-Flow layers. 
As indicated in rows $l=3,4$ and $l=2,3,4$, it is observed that while the metrics do not see a significant improvement, the number of trainable parameters increases substantially, \ie, each this layer causing about 5\% parameters. Consequently, we determine that placing a single Bi-Flow module at the third layer provides a better balance between tunable parameters and overall performance.

\subsubsection{Ablation on \texttt{X}-Adapter}
In this section, we ablate the \texttt{X}-Adapter structure under the emotion transfer task from four aspects, \ie, insert location, condition integration method, scale form, and PEFT variants. 
The results are summarized in Tab.~\ref{tab:abla}(b), where each row only contains one modification over the full version.

\noindent\textbf{Analysis on Insertion Form and Position (Tab.~\ref{tab:abla}(b) Location part).}
We explore how to insert the added adapter into the original network by comparing the parallel and sequential instances.
As shown in rows Serial and Parallel, the parallel form outperforms the sequential one in all metrics. This could be attributed to the parallel adapter receiving the same input as the sub-layers and directly updating the output, which is a more intuitive and natural design. 
This approach minimally impacts the original model, as the adapter operates independently of the sub-layer outputs. 
Moreover, we attempt to reduce the use of adapters to reduce the number of trainable parameters. 
However, from rows Only MHA and Only FFN, we observe that fine-tuning only the adapters inserted into either the FFN or MHA leads to a certain degree of performance decline under the same training settings. 
Thus, we incorporate our \texttt{X}-Adapter for \emph{both} the FFN and MHA.

\noindent\textbf{Analysis on Condition (Tab.~\ref{tab:abla}(b) Condition part).}
It is essential to incorporate the corresponding conditions to control the fine-tuning of emotions or identities. 
As shown in this part, when the injection of conditions is removed (row w/o. \texttt{X}), all metrics exhibit degradation, indicating that conditions can guide the network to better express the desired information. 
Moreover, the absence of conditions can also lead to a loss of editing capabilities during the inference phase. 
We further explore how to integrate the conditions. 
An intuitive approach is to use a stylization method~\cite{huang2017arbitrary}, mapping the conditions into the scale and shift factors to affect the original features, as shown in row Stylization. 
Although introducing this module enhances performance, it significantly increases the number of trainable parameters, \ie, from 4M to 8.14M. 
To balance the performance and cost, we continue to experiment with an extreme case, directly adding the condition to the original features without adding any trainable parameters. 
To our delight, we find that this simple method can achieve comparable results. Therefore, we employ the addition manner to exert the influence of conditions.

\begin{figure}[t!]
	\centering
	\includegraphics[width=0.45\textwidth]{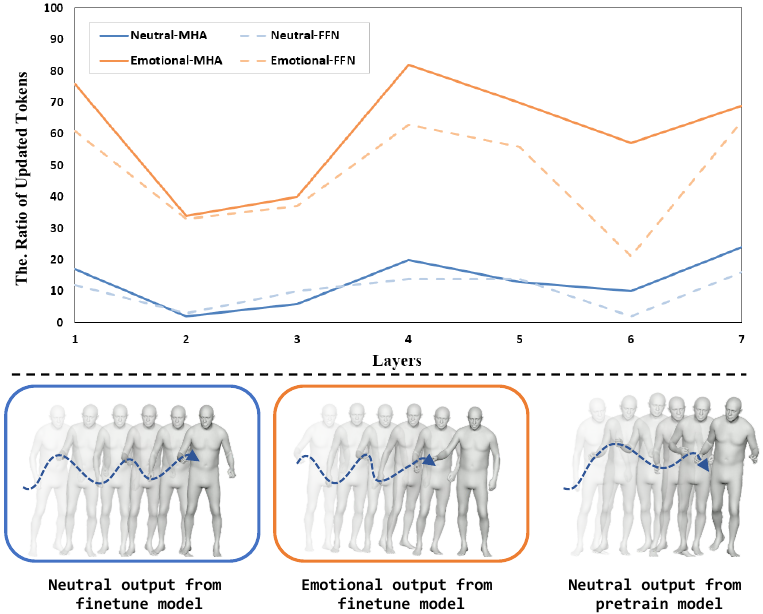}
	\caption{\textbf{Qualitative visualizations for Dy-Scale.} We display the ratio of updated tokens in Dy-Scale for each layer across different inputs (Neutral and Emotional), and the output motions of three variants.}
	\label{fig:abla_dy}
\end{figure}

\noindent\textbf{Analysis on Scale Form (Tab.~\ref{tab:abla}(b) Scale part).}
We introduce a dynamic scale mechanism (Dy-Scale) in \texttt{X}-Adapter to dynamically adjust the original features
by considering the significance score of the tunable features. 
To verify its effectiveness, we conduct experiments on two ablated variations: the fixed scalar scale which is set to 1.0 (row Scalar-1.0), and the learnable scalar scale which is initialized by 1.0 (row L-Scalar). 
As shown in this part, our Dy-Scale yields the best performance with a negligible increase in tunable parameters, from 3.99M to 4.00M. 
Thus, for such complex temporal motions, learning an adaptive weight for each frame is an intuitive and effective strategy. 
We further visualize the ratio of updated tokens in each layer for both MHA and FFN to explore the mechanism of Dy-Scale. 
As shown in Fig.~\ref{fig:abla_dy}, the \emph{same} audio input exhibits similar changes across different layers of MHA and FFN under various emotional conditions, but the specific ratios are entirely distinct. 
The neutral output from the fine-tuned model has only a few activated tokens because it shares similar motion patterns as the neutral output from the pre-trained model. 
In contrast, the emotional output is entirely different, which can also be discerned from the adjustment ratio. 

\noindent\textbf{Analysis on Other PEFT Methods (Tab.~\ref{tab:abla}(b) PEFT part).}
To further prove the effectiveness of our proposed \texttt{X}-Adapter, we compare it with several PEFT approaches, \ie, Prefix Tuning~\cite{li2021prefix} and LoRA~\cite{hu2021lora}, under the same training time and data. 
Specifically, prefix tuning involves prepending 64 learnable tokens in the temporal dimension and adding the conditions to them. 
From row Prefix Tuning, it is evident that this manner does not achieve effective transfer. 
While the number of trainable parameters is minimal, it cannot be ignored that an increase in the number of tokens also leads to a substantial increase in memory consumption. Besides, to ensure a fair comparison, we adapt Dy-Scale and the condition in the same manner as LoRA. 
We set the rank $r$ for both to be 64 thus the tunable parameters are the same. 
By comparing rows LoRA-r64 and Adapter-r64, the overall performance of the \texttt{X}-Adapter is superior to that of LoRA. 
Furthermore, we attempt to increase the rank to 128. 
As observed in rows Adapter-r128 and Adapter-r64, while the number of trainable parameters has increased, there is a noticeable improvement in overall performance. 
The potential reason may be that complex motion patterns require more parameters to be well-fitted during the fine-tuning phase. Experimentally, we employ the proposed \texttt{X}-Adapter with rank set to 128 in the full version.

\begin{table}[t!]
\caption{\textbf{Average scores of user study with 95\% confidence intervals.} The left part is for DU-Trans and the right is for \texttt{X}-Adapter.}
  \centering
  \scriptsize
   \renewcommand\arraystretch{1.2}
   \setlength\tabcolsep{5pt}
   \begin{tabular}
   {C{18pt}C{46pt}C{36pt}C{32pt}|C{18pt}C{41pt}}
      \toprule
         & Human-likeness  & Synchronized & Coordinated &  & Generalized-E   \\
      \midrule
      Exp 1 & 30.1$\pm$5.1  & 55.6$\pm$3.5 & 58.3$\pm$3.1 & Exp 5 & 55.8$\pm$3.2 \\
      Exp 2 & 50.0$\pm$3.9 & 56.0$\pm$2.5 & 58.9$\pm$4.7 & Exp 6 & 57.7$\pm$2.5\\
      Exp 3 & 69.9$\pm$5.0  & 63.2$\pm$3.3 & 67.4$\pm$3.5 & Exp 7 & 62.4$\pm$3.9 \\
      Exp 4 & 0.0+0.9 & - & - & Exp 8 & 61.9$\pm$4.3\\
      Ours & 84.6$\pm$4.3 & 72.2$\pm$4.0  & 75.1$\pm$4.0 & Ours & 72.7$\pm$5.1 \\
      \bottomrule
   \end{tabular}
   \label{tab:user2}
\end{table}

\subsubsection{User Study}
\label{sec:4.4.3}
We further conduct the user study of ablation studies as additional validation for the aforementioned objective conclusions. Similarly, we use the same approach as in Sec.~\ref{sec:4.2.3}.

For ablation on DU-Trans, we select four variants, Exp 1 corresponds to row $l=0, j=8$, Exp 2 to row $l=8, j=0$, Exp 3 to row $l=7, j=1$, and Exp 4 to row w/o. $\lambda_F,\lambda_B$, respectively, while Ours corresponds to row $l=3$ in Tab.~\ref{tab:abla}(a). We adopt the human-likeness (20 combinations appear once, 20 video pairs, 40 qualified participants), synchronized (8 video pairs for each method, 37 qualified participants), and coordinated (8 video pairs for each method, 38 qualified participants) three user studies. The average score shown in the left part of Tab.~\ref{tab:user2}. Notably, we do not include the mismatching tests for Exp 4 since it fails to produce plausible human motions, as evidenced in Fig.~\ref{fig:abla} row 4. Specifically, comparing Exps 1–3 shows that our divided encoder and unified decoder preserve individual differences while achieving commonality. Moreover, comparing Exps 3 and 4 reveals that the auxiliary losses for face and body help capture their unique features. Lastly, comparing Exp 3 and Ours confirms that the Bi-Flow mechanism further boosts performance.

For ablation on \texttt{X}-Adapter, to align with the Tab.~\ref{tab:abla}, we only evaluate emotion transfer capability through the Generalized-E test (8 video pairs for each method, 38 qualified participants). The average score shown in the right part of Tab.~\ref{tab:user2}. Exp 5 corresponds to row Serial, Exp 6 to row w/o. \texttt{X}, Exp 7 to row L-Scalar, and Exp 8 to row Adapter-r64, while Ours corresponds to row Adapter-r128 in Tab.~\ref{tab:abla}(b). Our method's mismatched percentage is higher than that of the four variants. This indicates that: 1) The parallel insertion method outperforms the serial one. 2) The absence of an emotional condition significantly degrades transfer accuracy. 3) Our proposed dynamic learning with an adaptive weight per frame (Dy-Scale) is superior to using a global weight (L-Scalar). 4) Increasing the adapter's rank enhances performance, respectively.

\section{Conclusion}
In this work, we focus on enhancing the user experience with talking avatars, concentrating on the harmony of full-body movements and the rapid adaptation of new identity and emotional data. 
To achieve such, we propose $\cb$, which includes two critical designs: 
1) DU-Trans operates by initially dividing into dual pathways designed to independently learn the distinct features of the face and body, each guided by auxiliary losses and enriched with holistic dynamic priors through the Bi-Flow layer. Then, it unites the learned two features to model a joint distribution and directly predicts the combined coefficients that ensure a high degree of coordinated holistic motions. 
2) \texttt{X}-Adapter seamlessly integrates into a pretrained DU-Trans network that can quickly transfer the original model to an emotional one or other totally different identities with much fewer trainable parameters. 
Our approach has demonstrated state-of-the-art performance on two public datasets, along with an efficient ability to transfer identities and emotions.

\noindent\textbf{Limitations: } Despite the significant improvements, $\cb$ still suffers from several limitations. 
First, since we do not perform targeted design for the foot trajectory, the motion generated by our method exhibits physical implausibilities of foot sliding. 
We plan to introduce explicit physical modeling to mitigate this issue. 
Second, due to the limitations of BEAT2 and SHOW datasets, we are unable to generate highly photo-realistic avatars, which is crucial for enhancing the user experience. 
We aim to develop a large-scale multi-view full-body talking avatar dataset to prompt the advancement of this field.

\section*{Acknowledgements}
This work is supported in part by National Science and Technology Major Project under contract No. 2021ZD0109902, in part by the National Natural Science Foundation of China under contract No. 62171256.

\ifCLASSOPTIONcaptionsoff
  \newpage
\fi

\bibliographystyle{IEEEtran}
\bibliography{reference}

\end{document}